\documentclass{article}

\usepackage{microtype}
\usepackage{graphicx}
\usepackage{subcaption}
\usepackage{booktabs} 

\usepackage{hyperref}

\usepackage[preprint]{icml2026}

\usepackage{amsmath}
\usepackage{amssymb}
\usepackage{mathtools}
\usepackage{amsthm}

\usepackage{booktabs}       
\usepackage{xcolor}         
\usepackage{graphicx}
\usepackage{multirow}
\usepackage{multicol}
\usepackage{algorithm}
\usepackage{algorithmic}
\usepackage{wrapfig}
\usepackage[most]{tcolorbox} 
\DeclareMathOperator*{\argmax}{arg\,max}

\usepackage[capitalize,noabbrev]{cleveref}

\theoremstyle{plain}

\theoremstyle{definition}

\theoremstyle{remark}

\usepackage[textsize=tiny]{todonotes}

\icmltitlerunning{Finding the Cracks: Improving LLMs Reasoning with Paraphrastic Probing and Consistency Verification}

\begin{document}

\twocolumn[
  \icmltitle{Finding the Cracks: Improving LLMs Reasoning with Paraphrastic Probing and Consistency Verification}



  \icmlsetsymbol{equal}{*}

  \begin{icmlauthorlist}
    \icmlauthor{Weili Shi}{yyy}
    \icmlauthor{Dongliang Guo}{yyy}
    \icmlauthor{Lehan Yang}{yyy}
    \icmlauthor{Tianlong Wang}{yyy}
    \icmlauthor{Hanzhang Yuan}{yyy}
    \icmlauthor{Sheng Li}{yyy}
  \end{icmlauthorlist}

  \icmlaffiliation{yyy}{School of Data Science, University of Virginia, Charlottesville, United States.}

  \icmlcorrespondingauthor{Weili Shi}{weili.shi@virginia.edu}
  \icmlcorrespondingauthor{Sheng Li}{sheng.li@virginia.edu}

  \icmlkeywords{Machine Learning, ICML}

  \vskip 0.3in
]



\printAffiliationsAndNotice{}  

\begin{abstract}
Large language models have demonstrated impressive performance across a variety of reasoning tasks. However, their problem-solving ability often declines on more complex tasks due to hallucinations and the accumulation of errors within these intermediate steps. Recent work has introduced the notion of \textit{critical tokens}--tokens in the reasoning process that exert significant influence on subsequent steps. Prior studies suggest that replacing critical tokens can refine reasoning trajectories. Nonetheless, reliably identifying and exploiting critical tokens remains challenging. To address this, we propose the \textbf{P}araphrastic \textbf{P}robing and \textbf{C}onsistency \textbf{V}erification~(\textbf{PPCV}) framework. PPCV operates in two stages. In the first stage, we roll out an initial reasoning path from the original question and then concatenate paraphrased versions of the question with this reasoning path. And we identify critical tokens based on mismatches between the predicted top-1 token and the expected token in the reasoning path. A criterion is employed to confirm the final critical token. In the second stage, we substitute critical tokens with candidate alternatives and roll out new reasoning paths for both the original and paraphrased questions. The final answer is determined by checking the consistency of outputs across these parallel reasoning processes. We evaluate PPCV on mainstream LLMs across multiple benchmarks. Extensive experiments demonstrate PPCV substantially enhances the reasoning performance of LLMs compared to baselines.
\end{abstract}

\begin{figure}[t]
    \centerline{\includegraphics[width=0.85\columnwidth]{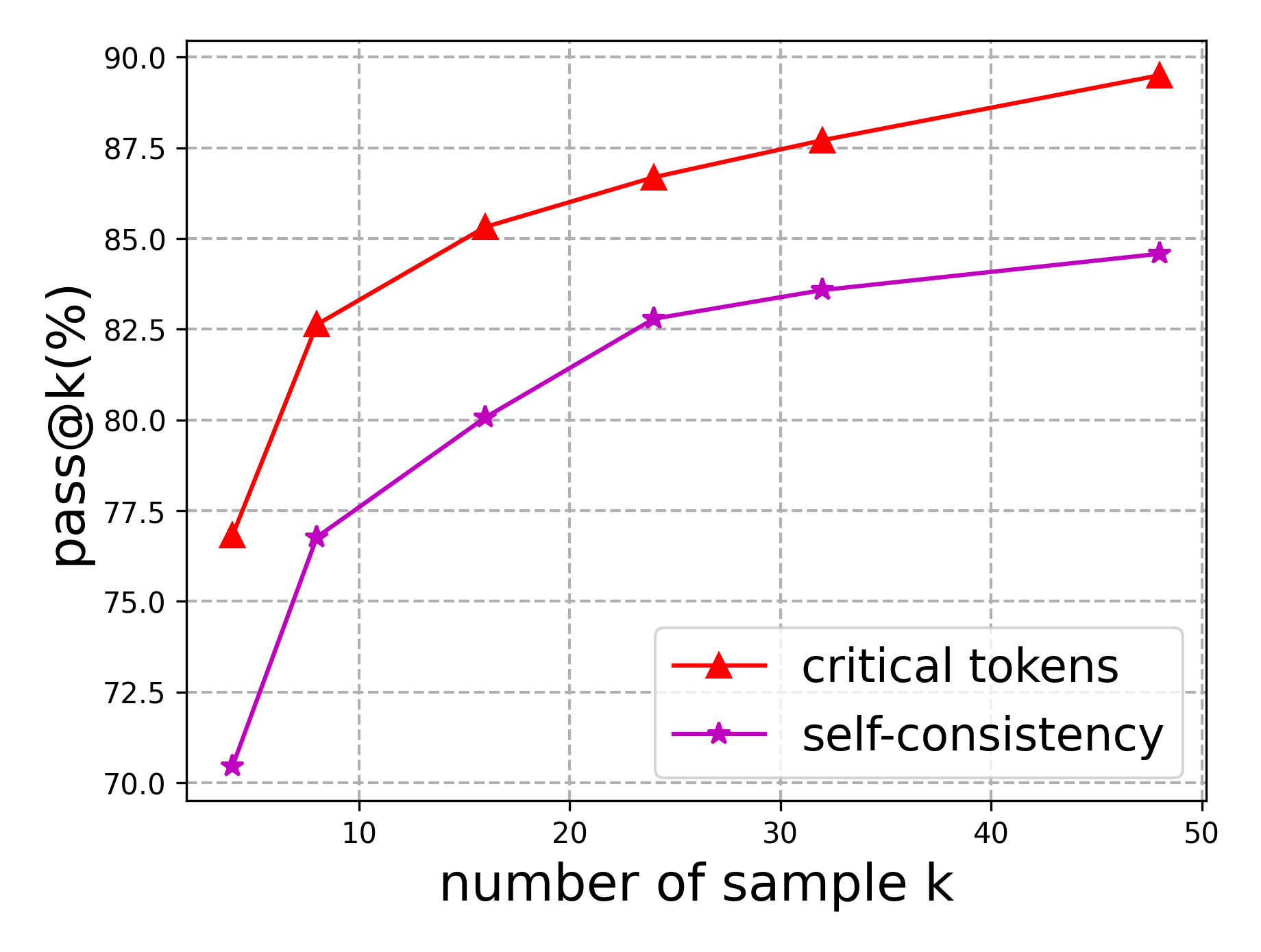}}
    \vspace{-2mm}
    \caption{Comparison of the effects of critical tokens and Self-Consistency on the reasoning performance of LLMs, evaluated on samples from the GSM8K training data.}
    \label{fig:compare_ct_sc_gsm8k}
    \vspace{-5mm}
\end{figure}

\section{Introduction}

The emergence of large language models~(LLMs)~\citep{brown2020language,grattafiori2024llama,achiam2023gpt,yang2025qwen3} has astonished the AI community with their remarkable capabilities across a wide range of reasoning tasks, including mathematical problem solving, programming, and planning. By generating intermediate reasoning steps through techniques such as chain-of-thought prompting~\citep{wei2022chain,kojima2022large,zhangautomatic,spraguecot}, LLMs can emulate human-like reasoning processes and achieve strong performance on diverse reasoning benchmarks. Despite their success, the problem-solving ability of LLMs often declines on complex reasoning tasks due to hallucinations and the accumulation of errors in intermediate steps~\citep{lightman2023let,ling2023deductive,bubeck2023sparks}. To mitigate this issue, prior works~\citep{wang2024math,yuanfree,chenboosting,chentoward} have proposed the inference-time optimization paradigm, which evaluates and refines reasoning trajectories through self-correction by the LLM itself or with feedback from external verifiers such as process reward models~(PRMs)~\citep{uesato2022solving}. However, existing studies~\citep{yangsupercorrect,tyen2024llms,stechlyself} suggest that LLMs struggle to reliably identify errors in their reasoning and often fail to correct previous outputs without external guidance. Moreover, obtaining high-quality, stepwise process supervision for training external verifiers is challenging and limits the practicality of these approaches~\citep{fengstep}.

\begin{figure}[t]
    \centering
    \includegraphics[width=1.00\columnwidth]{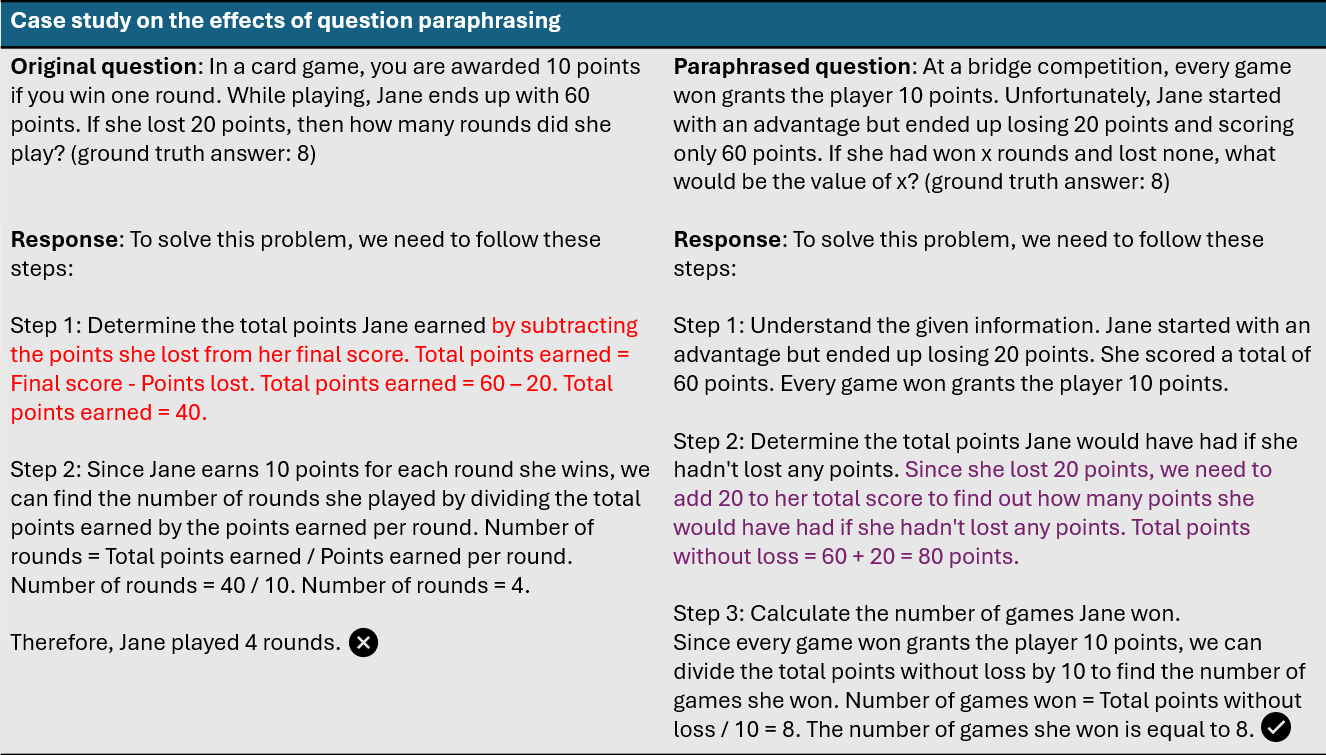}
    \caption{Case study illustrating that LLMs can refine intermediate incorrect reasoning segments~(highlighted in red).}
    \vspace{-3mm}
    \label{fig:para_example}
\end{figure}

\begin{figure}[t]
    \centerline{\includegraphics[width=1.00\columnwidth]{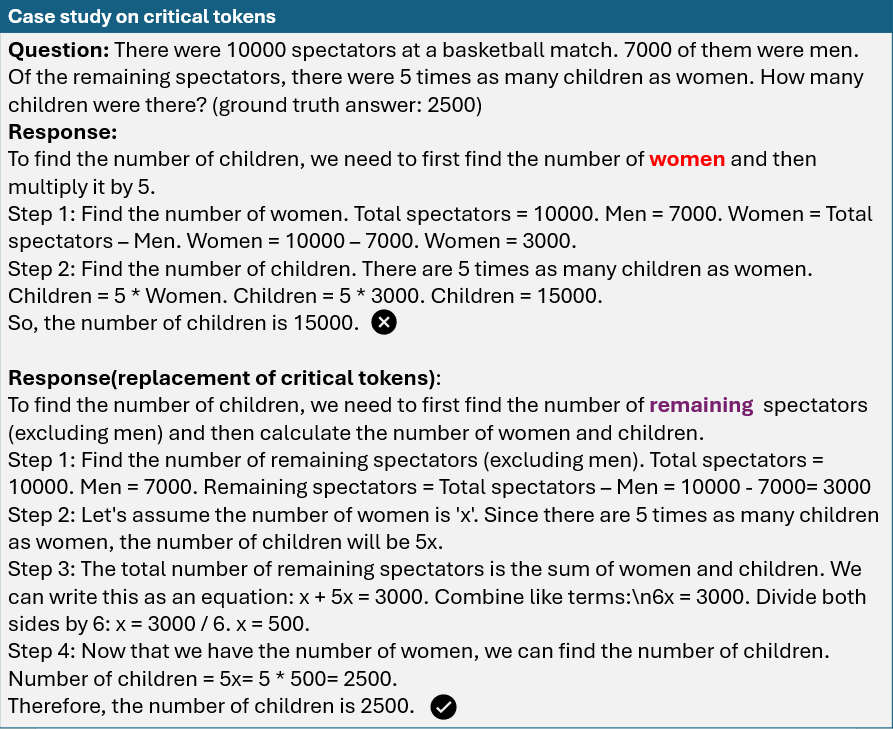}}
    \caption{An example demonstrating how substitution of a critical token (red) with a candidate token (purple) modifies the reasoning path and produces the correct answer.}
    \label{fig:critical_token_example}
    \vspace{-5mm}
\end{figure}

Recent work has introduced the concept of \textit{critical tokens}~\citep{lincritical}, which play pivotal roles in intermediate reasoning steps and exert strong influence on subsequent reasoning and final outcomes. We evaluate the criticality of a token based on the average accuracy of sampling rollouts generated from the partial reasoning trajectories truncated at the token’s position. A token is identified as critical if it lies on an incorrect trajectory and satisfies the following conditions:
\vspace{-4mm}
\begin{itemize}
    \item The correctness score of rollouts initiated from the token is close to zero (e.g., below 5\%);
    \vspace{-3mm}
    \item The correctness scores of rollouts initiated from subsequent tokens remain below a threshold (e.g., 10\%).
\end{itemize}
\vspace{-3mm}
Prior studies suggest that critical tokens often diverge from human-annotated error tokens, yet they induce more sustained degradation in reasoning quality than other tokens. Moreover, as illustrated in Figure~\ref{fig:critical_token_example}, replacing critical tokens in an incorrect reasoning trajectory with suitable candidate tokens can correct subsequent steps and lead to the right answer in new rollouts. To quantitatively assess the effectiveness of critical tokens, we conduct an empirical study using LLMs such as Llama-3.1-8B-Instruct~\citep{grattafiori2024llama} on reasoning tasks. Specifically, we randomly sample 100 instances with incorrect reasoning steps from the GSM8K~\citep{cobbe2021training}  training data. Following the criterion, we locate critical tokens through exhaustive search. We then truncate the reasoning path at the critical token, substitute it with alternative tokens, and roll out new reasoning paths. For example, as shown in Figure~\ref{fig:critical_token_example}, the token “woman” is replaced with “remaining”. We evaluate performance using Pass@k and compare against Self-Consistency~\citep{wangself}, which also samples multiple reasoning paths. As shown in Figure~\ref{fig:compare_ct_sc_gsm8k}, critical token replacement provides a clear advantage in correcting errors compared to plain sampling. Nonetheless, reliably identifying and leveraging critical tokens for reasoning remains a nontrivial challenge. 

Recent studies~\citep{zhou2024paraphrase,chen2024self}, on \textit{surface form}, the way questions, assumptions, and constraints are phrased, have revealed its subtle influence on the trajectory of intermediate reasoning steps, As shown in Figure~\ref{fig:para_example}. LLMs could adjust the intermediate steps under the paraphrased form of the question. This motivates us to explore the role of paraphrasing in the extraction and utilization of critical tokens for reasoning tasks. To this end, we propose the \textbf{P}araphrastic \textbf{P}robing and \textbf{C}onsistency \textbf{V}erification~(\textbf{PPCV}) framework , a two-stage approach designed to leverage critical tokens to enhance the reasoning ability of LLMs. In the first stage, we probe critical tokens using paraphrased questions. Specifically, we first roll out the initial reasoning path from the original question, then concatenate paraphrased questions with this reasoning path. The resulting synthetic inputs are fed into the LLM to obtain token-level logits for each position in the reasoning path. Positions where the predicted top-1 token diverges from the expected token are flagged as potential pivotal points, as these positions are sensitive to paraphrased inputs and can trigger a pivot in the reasoning trajectory. Next, an empirical criterion is applied to determine the final critical token. In contrast to prior work~\citep{lincritical}, which depends on external models for identifying critical tokens with ambiguous criteria, our method introduces a self-contained mechanism that pinpoints critical tokens.

In the second stage, we leverage the extracted critical tokens to refine the initial reasoning path. Specifically, we select the top-K tokens~(include critical token itself) at the critical token position and roll out new reasoning paths for both the original and paraphrased questions. Based on the empirical observation that trajectories leading to correct answers are robust to paraphrastic perturbations, we propose a paraphrase consistency mechanism. In contrast to Self-Consistency~\citep{wangself}, which relies on majority voting, our method selects the final answer by comparing outputs from paraphrased and original questions and choosing the one with the most consistent matches. To address potential ties across multiple answers, we further introduce similarity-weighted paraphrase consistency, which incorporates similarity scores between paraphrased and original questions when computing consistency.

Compared with self-correction~\citep{wu2024large,miaoselfcheck} and PRM-based methods~\citep{wang2024math,yuanfree}, our framework exploits critical tokens to refine reasoning trajectories without requiring step-level error detection by the LLM itself or auxiliary models. We evaluate our method on mainstream LLMs across mathematical and commonsense reasoning benchmarks, demonstrating consistent improvements in reasoning performance. The contributions of the paper is summarized as follows:
\begin{itemize}
    \item We propose a novel two-stage framework,  \textbf{P}araphrastic \textbf{P}robing and \textbf{C}onsistency \textbf{V}erification~(\textbf{PPCV}) designed to extract and leverage critical tokens to enhance the reasoning performance of LLMs.
    \item We show that critical tokens can more effectively correct erroneous reasoning trajectories than traditional sampling methods like Self-Consistency. Furthermore, our approach successfully extracts these tokens through paraphrastic probing, achieving improved final results via paraphrase consistency.
    \item We evaluate our method on mainstream LLMs across various reasoning tasks, including math and logic reasoning. Experimental results show significant performance improvements over baseline methods.
\end{itemize}

\section{Related Work}
\textbf{Inference-Time Optimization for LLM reasoning.} With the advent of chain-of-thought~(CoT) prompting, LLMs have demonstrated strong reasoning capabilities by producing intermediate steps during inference. This success has motivated a growing body of work~\citep{wu2025inference,snell2024scaling} on augmenting reasoning trajectories at test time to further improve performance. Existing approaches can be broadly categorized into search-based methods~\citep{biforest,yao2023tree,hao2023reasoning,xie2023self,besta2024graph}, such as Tree-of-Thoughts~\citep{yao2023tree}, and sampling-based methods~\citep{wangself,xu2025phi,wan2025reasoning,manon}, such as Self-Consistency~\citep{wangself}. However, due to intrinsic hallucinations~\citep{bubeck2023sparks}, LLMs often generate erroneous intermediate steps, which can ultimately lead to incorrect answers, especially on complex problems. This limitation highlights the need for inference-time optimization of reasoning processes.

To address this issue, one line of research~\citep{yin2024reasoning,chenboosting,ling2023deductive,wu2024large,miaoselfcheck,madaan2023self} designs instructional prompts that guide LLMs to detect and refine their own mistakes. Despite its appeal, prior work has shown that the effectiveness of self-correction is limited in practice. Another line of work~\citep{wang2024math,yuanfree,he2024advancing,havrillaglore} introduces external verifiers, such as process reward models~\citep{snell2024scaling}, to identify and filter out error-prone reasoning steps. These methods typically require high-quality training data for the verifier, with data scarcity often mitigated through strategies such as Monte Carlo Tree Search~\cite{guanrstar,qimutual,lipolicy,zhang2024rest}. In addition, a recent line of decoding-based approaches~\citep{xu2025phi,manon} seeks to improve reasoning by dynamically adjusting the next-token prediction based on future trajectory probing. In contrast, our method refines reasoning by directly leveraging critical tokens, without relying on stepwise verification or external verifiers. This design underscores both the utility and universality of our approach.

\textbf{Paraphrasing for LLMs.} A growing number of work~\citep{zhou2024paraphrase,chen2024self} has examined the impact of a problem’s surface form on the reasoning ability of LLMs. Findings~\citep{zhou2024paraphrase,chen2024self,huangmath} suggest that even subtle modifications in phrasing can substantially affect both the reasoning process and the final outcome. Building on this observation, several methods~\citep{yadav2024pag,chen2024self} leverage beneficial paraphrasing to enhance LLM performance in tasks such as reasoning and intent classification. In addition, paraphrasing has been employed to assess model uncertainty~\citep{gao2024spuq,tanneru2024quantifying}, thereby enhancing the reliability and trustworthiness of LLM applications. In our work, we utilize paraphrasing as a principled tool to extract critical tokens and to aggregate answers.

\begin{figure*}[ht]
    \centering
    \includegraphics[width=0.95\textwidth]{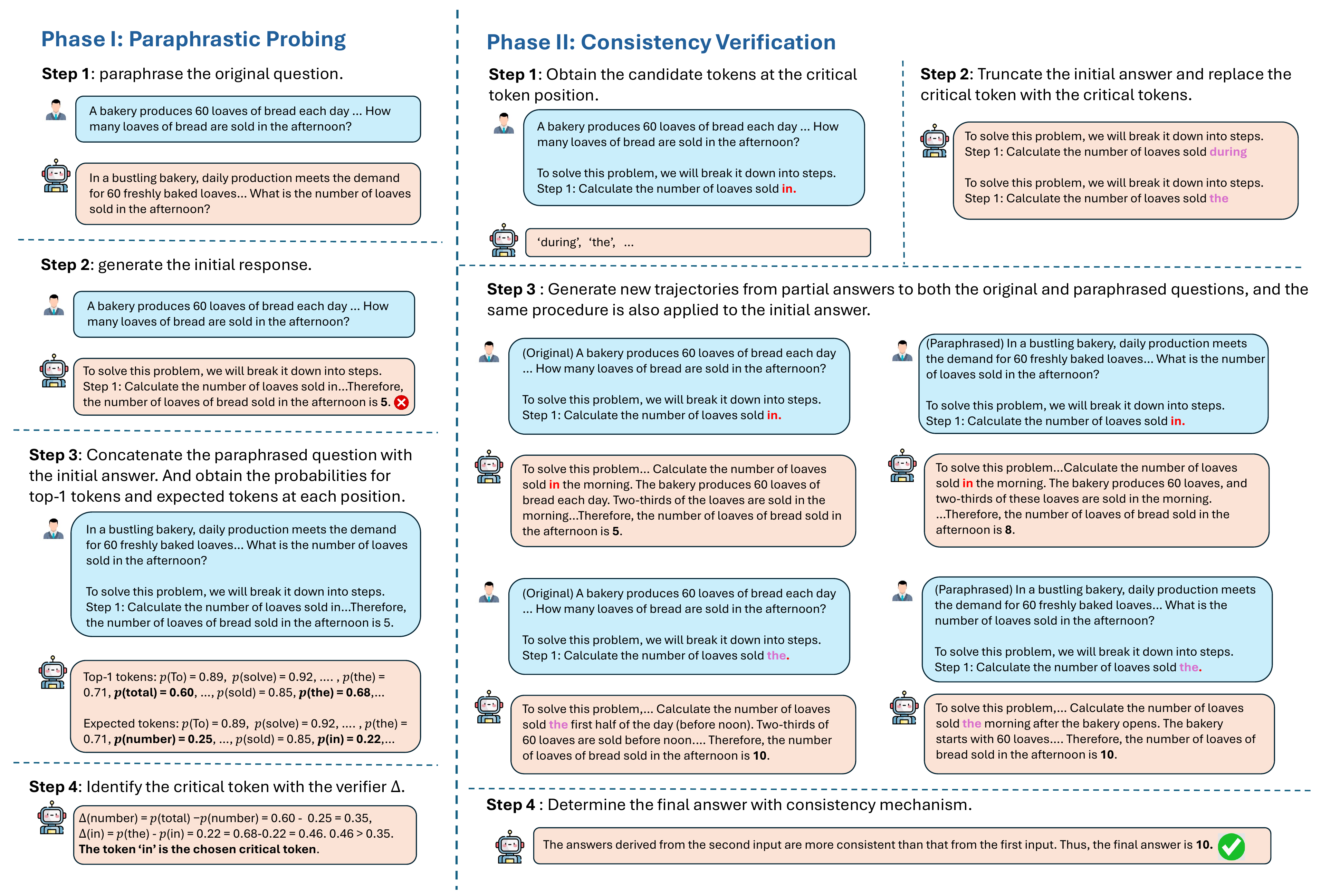}
    \vspace{-3mm}
    \caption{Illustration of the proposed paraphrastic probing and consistency verification~(PPCV) framework. The framework comprises two stages: (i) probing critical tokens through paraphrased forms, and (ii) rolling out new reasoning steps with alternative tokens and selecting the final answer using the paraphrase consistency verification mechanism. }
    \vspace{-3mm}
    \label{fig:framework}
\end{figure*}

\begin{figure}[ht]
    \centering
    \includegraphics[width=0.95\columnwidth]{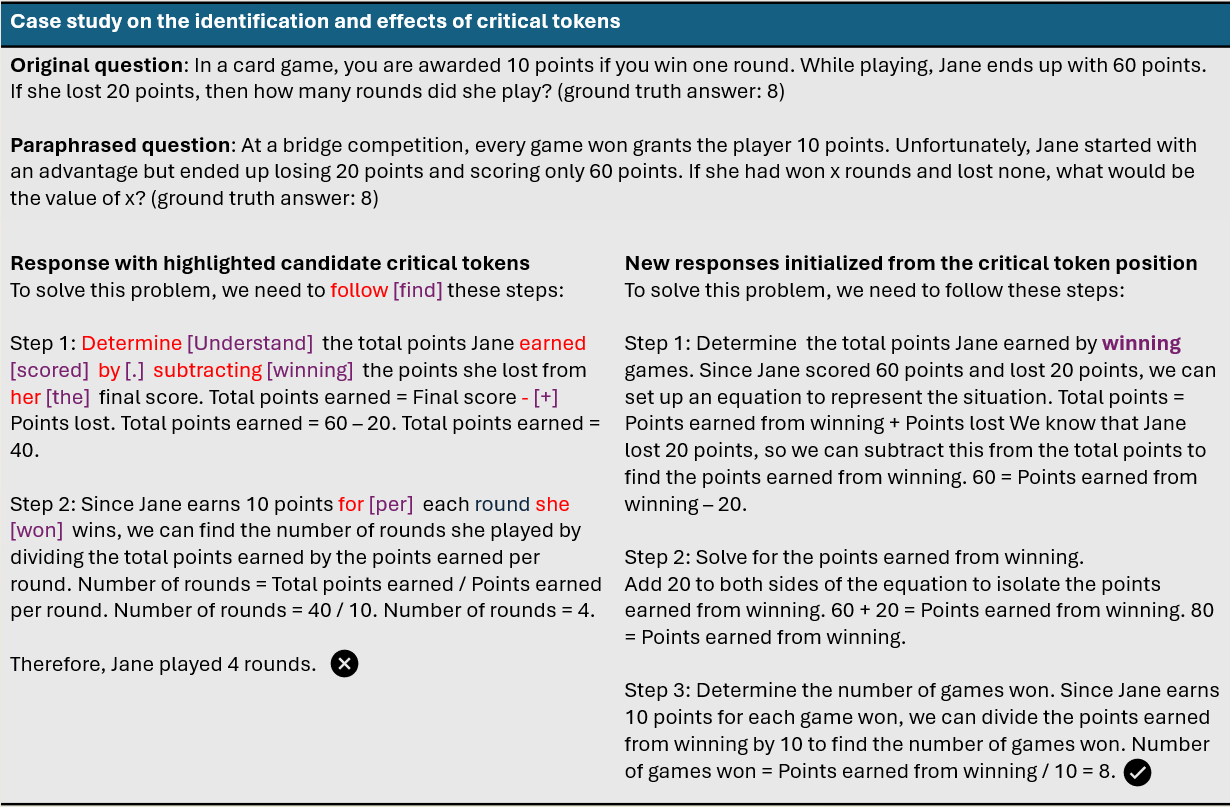}
    \caption{Case study illustrating the identification and effects of critical tokens identified by our method. Tokens highlighted in red indicate candidate critical tokens, whereas tokens highlighted in purple correspond to alternative tokens generated when conditioning on paraphrased questions.}
    \vspace{-3mm}
    \label{fig:probe_illustration}
    \vspace{-4mm}
\end{figure}

\section{Methodology}
In this section, we present the two components of our framework in detail: paraphrastic probing and consistency verification. An detailed illustration of our framework is shown in Figure~\ref{fig:framework}. We then discuss the proposed method and provide the complete algorithm.

\subsection{Paraphrastic Probing}
Previous findings~\citep{zhou2024paraphrase,chen2024self,huangmath} on the impact of a problem’s surface form suggest that the quality of intermediate reasoning steps is influenced not only by the underlying mathematical relationships and logic, but also by how the problem is expressed. Notably, LLMs are sometimes able to solve a paraphrased version of a problem that they fail to solve in its original form, highlighting the potential of paraphrasing to uncover pivotal tokens that are critical for successful reasoning. Motivated by this observation, we introduce paraphrastic probing to efficiently identify the critical token. Given the original question $q_0$, we first prompt the LLM to generate multiple paraphrased forms, denoted as $q_1,q_2,...,q_N$, where $N$ is the number of the paraphrased questions. We adopt Automatic Prompt Engineering~(APE)~\citep{zhou2022large} to derive paraphrasing instructions that preserve semantic integrity and all numerical values, mathematical relationships, and core logical structures of the problem, while maximizing linguistic and contextual diversity. Additional details can be found in Appendix~\ref{sec:app_ape}. We then obtain the initial reasoning path $r^{q_0}_0$ for the original question using greedy decoding. This reasoning path is subsequently concatenated with each paraphrased question, and the resulting synthetic inputs are fed into the LLM to compute the token probability distribution at each position in $r^{q_0}_0$. Specifically, the token probability distribution at $i$th position conditioned on the paraphrased question $q_n$ is expressed as
\begin{equation}
    P_i^{q_n} = \text{LLM}(\tilde{a}_i|\mathcal{I},q_n,r^{q_0}_{0,<i}),
    \label{eq:token_probability}
\end{equation}
where $\mathcal{I}$ denotes the instruction prefix and $\tilde{a}_i$ represents the sampled token at $i$th position. The token $\tilde{a}_i$ is regarded as a candidate critical token if predicted top-1 token does not match the expected token at the same position in $r^{q_0}_0$, i.e.,
\begin{equation}
    \argmax P_i^{q_n} \neq a_i,
    \label{eq:choose_candidate}
\end{equation}
where $a_i$ denotes the token at the $i$th position in $r^{q_0}_0$.

\begin{figure*}[t]
\centering
\subfloat[]{%
  \includegraphics[width=0.25\textwidth]{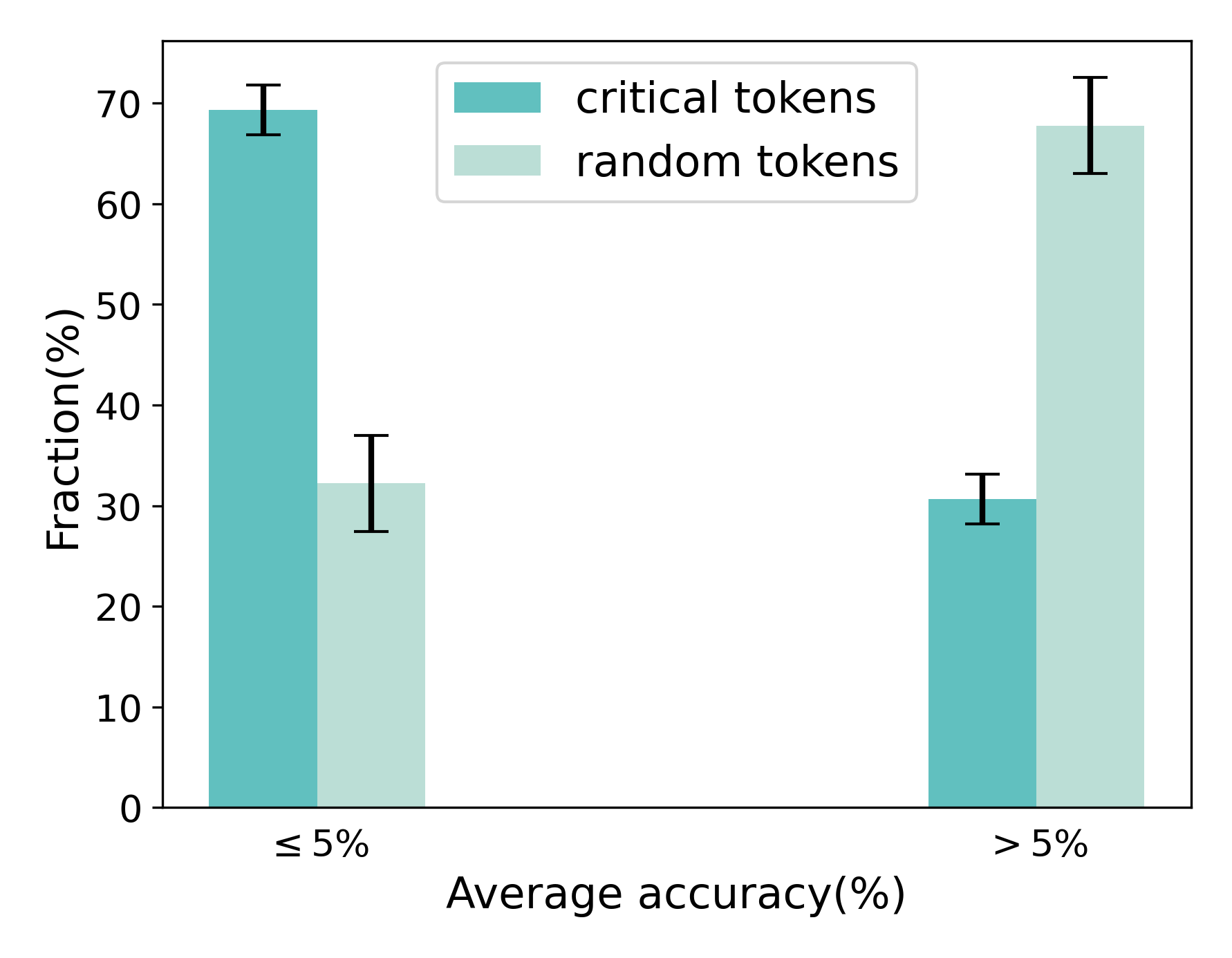}
  \label{fig:condition_one_gsm8k}
}
\subfloat[]{%
  \includegraphics[width=0.25\textwidth]{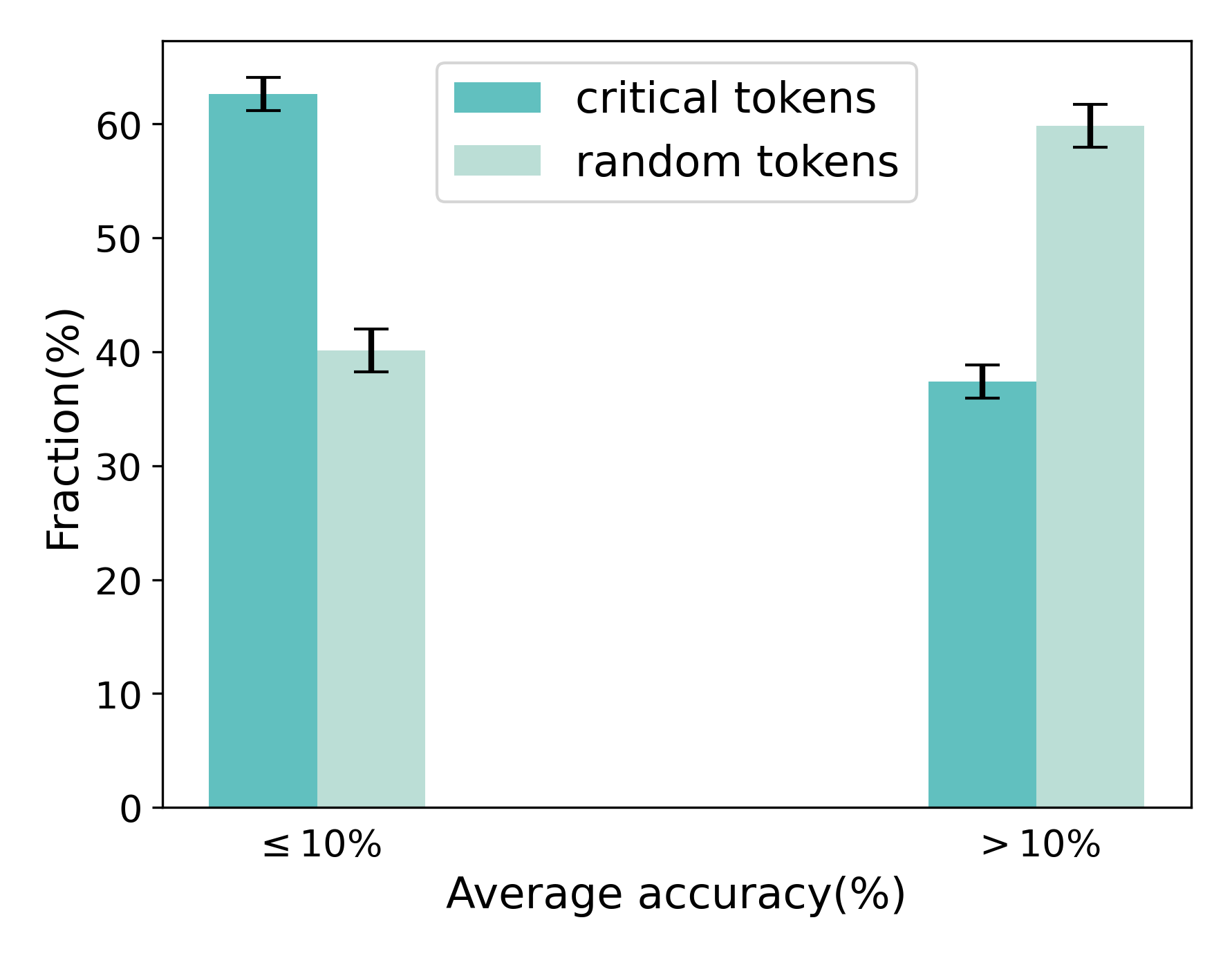}
  \label{fig:condition_two_gsm8k}
}
\subfloat[]{%
  \includegraphics[width=0.25\textwidth]{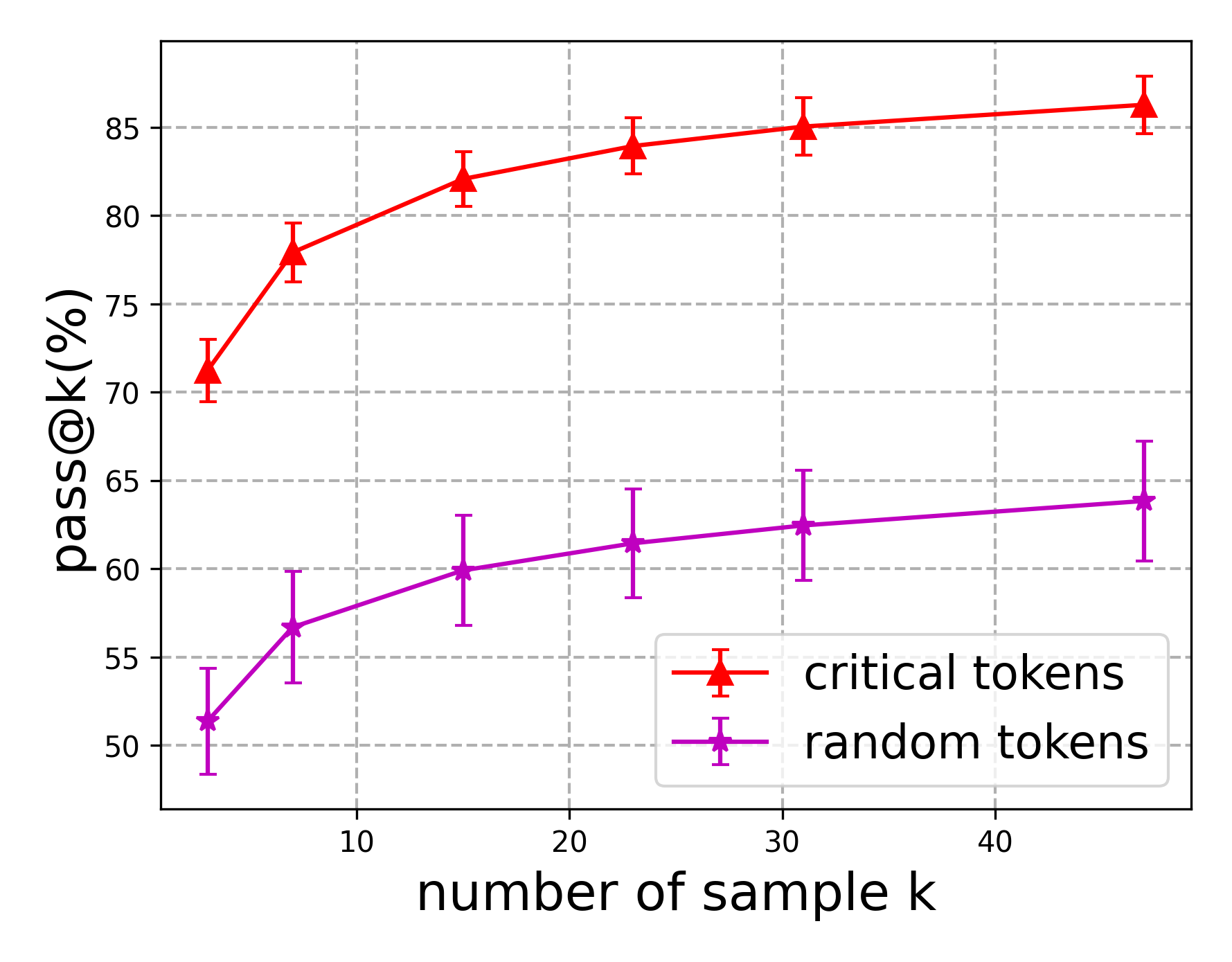}
  \label{fig:compare_ct_rnd_gsm8k}
}
\subfloat[]{%
  \includegraphics[width=0.25\textwidth]{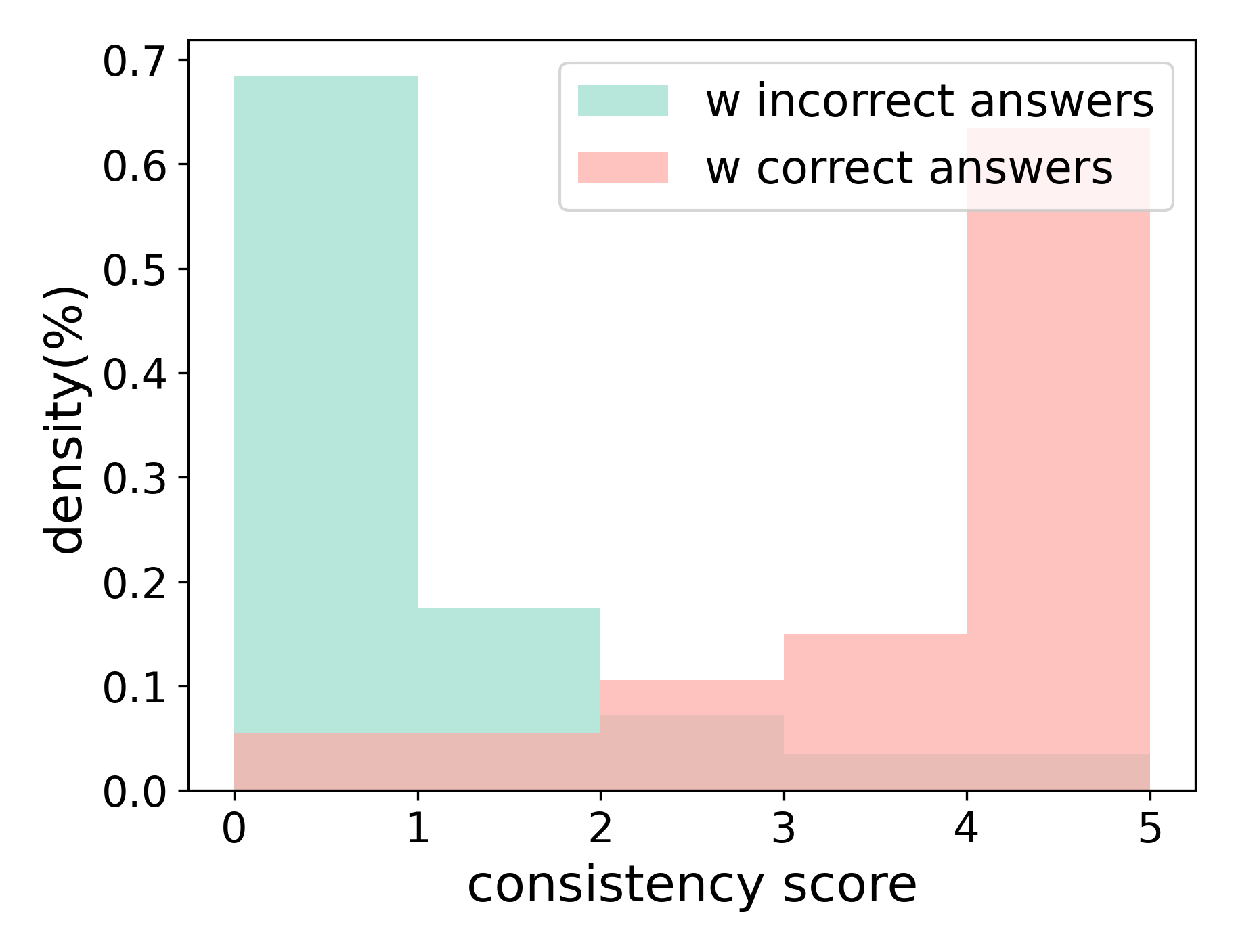}
  \label{fig:density_gsm8k}
}
\vspace{-2mm}
\caption{(a) Comparison of the fraction of tokens whose associated rollout correctness falls below or above 5\%, for tokens extracted by our method versus randomly selected tokens. (b) Comparison of the fraction of the subsequent tokens whose associated rollout correctness falls below or above 10\%, for tokens extracted by our method versus randomly selected tokens. (c) Comparison of the impact of extracted tokens by our method versus random tokens on LLM reasoning performance. (d) Comparison of the density distributions of consistency scores for rollouts with correct and incorrect answers.}
\vspace{-4mm}
\end{figure*}

To validate the effectiveness of our method in identifying critical tokens and to demonstrate the pivotal role these tokens play when extracted via paraphrastic probing, we conduct a case study illustrated in Figure~\ref{fig:probe_illustration}. Because large language models can adjust their reasoning trajectories under the paraphrased form of the question, certain tokens in the original reasoning sequence produce different subsequent tokens when conditioned on paraphrased inputs. In this example, most candidate critical tokens are located within incorrect reasoning segments. Notably, when we identify “subtracting” as a critical token and replace it with an alternative token (i.e., “winning”), the resulting rollout corrects the erroneous reasoning steps and yields the correct final answer. This example highlights the impact of critical tokens and underscores the effectiveness of our method in identifying tokens that are pivotal to reasoning outcomes.

In addition, we conduct a quantitative analysis to examine the authenticity and impact of tokens extracted via paraphrastic probing, comparing them against randomly selected tokens. Specifically, we sample 100 instances with incorrect reasoning trajectories from the GSM8K~\citep{cobbe2021training} training sets. Following the paraphrastic probing pipeline, we identify candidate critical tokens in the early reasoning steps. In each run, we randomly sample 40\% of the candidate critical tokens for evaluation and repeat the experiment independently 10 times. For comparison, we apply the same procedure to randomly selected tokens. All evaluations are conducted using Llama-3.1-8B-Instruct~\citep{grattafiori2024llama}. We first compute the average accuracy of rollouts generated from partial trajectories truncated at the position of the extracted tokens, and compare these results with those obtained from random-token truncation. As shown in Figure~\ref{fig:condition_one_gsm8k}, a large proportion of the extracted tokens exhibit near-zero correctness, consistent with the first criterion of critical tokens. We further evaluate the average accuracy of rollouts initiated from subsequent tokens (5–10 tokens). The results in Figure~\ref{fig:condition_two_gsm8k} indicate that errors persist beyond the identified positions, supporting the second criterion that critical tokens induce sustained degradation in downstream reasoning. Finally, we substitute the extracted critical tokens with alternative tokens and roll out multiple new reasoning trajectories. As shown in Figure~\ref{fig:compare_ct_rnd_gsm8k}, replacing critical tokens leads to a significantly larger improvement in reasoning accuracy compared to substituting random tokens. These results further validate both the pivotal role of the identified critical tokens and the effectiveness of our paraphrastic probing method. Additional results can be found in Appendix~\ref{sec:app_empirical}.

We introduce a heuristic verifier to select the final critical token from multiple candidates. For a candidate token $a_i$ and paraphrased question $q_n$, the verification score is defined as
\begin{equation}
    \Delta_{q_n} (a_i) = \max P^{q_n}_i-P^{q_n}_i(\tilde{a}_i=a_i).
    \label{eq:criterion_1}
\end{equation}
where $P^{q_n}_i$ denotes the predictive distribution at position $i$ on question $q_n$. Intuitively, $\Delta$ measures how much the predicted top-1 token deviates from the expected token, indicating the token’s potential impact on the reasoning trajectory. For each extracted token $a_i$,we take the maximum score across paraphrases,
\begin{equation}
    \Delta (a_i) = \max_{q_n} \Delta_{q_n} (a_i),
    \label{eq:criterion_2}
    \vspace{-2mm}
\end{equation}
and select the final critical token as
\vspace{-2mm}
\begin{equation}
   a_c = \argmax_i \Delta (a_i).
   \label{eq:criterion_3}
   \vspace{-2mm}
\end{equation}

\subsection{Consistency Verification}
After identifying the final critical token $a_c$, we aim to refine the original reasoning path with alternative tokens and achieve final answer with paraphrase consistency mechanism. Specifically,  we generate a set of alternative tokens $a^0_c, a^1_c, a^2_c,..., a^{K-1}_c$ using the LLM conditioned on original question $q_0$, where $a^0_c$ is the original token in $r^{q_0}_0$ and the remaining tokens are sampled via top-K sampling. The initial reasoning path is truncated at the position of critical token, and each alternative token is concatenated to form synthetic inputs $\tilde{r}_c^0, \tilde{r}_c^1, \tilde{r}_c^2,...,\tilde{r}_c^{K-1}$. We then roll out new reasoning trajectories for each synthetic input with respect to both the original and paraphrased questions using greedy decoding, denoted as $r^{q_0}_k,r^{q_1}_k,...,r^{q_N}_k$ for $k=0, 1, 2,...,K-1$. Next, for the rollout with the $k$th alternative token, we compare the answers obtained from the paraphrased forms with that from the original form and compute a consistency score  $c_k = \sum_{n-1}^N \mathbb{I}(\Phi(r^{q_0}_k)=\Phi(r^{q_n}_k))$, where $\Phi(\cdot)$ and $\mathbb{I}(\cdot)$ denotes the function that extracts the final answer from a reasoning trajectory and the indicator function, respectively. The answer associated with the highest consistency score is then selected as the final prediction
\begin{equation}
    \text{ans}_{f} = \Phi(r^{q_0}_k), \text{where} \, k = \argmax_k c_k.
    \label{eq:consistency}
    \vspace{-4mm}
\end{equation}

To justify our paraphrase consistency mechanism, we investigate the impact of paraphrased forms on LLM reasoning. We sample instances from the GSM8K~\citep{cobbe2021training} and follow our pipeline to extract critical tokens. From each truncated reasoning trajectory, we roll out multiple reasoning paths by concatenating alternative tokens. For each original question, we generate five paraphrased variants and compute the consistency score for resulting rollouts. The evaluation is conducted on Llama-3.1-8B-Instruct~\citep{grattafiori2024llama}. We then analyze the distribution of consistency scores for rollouts that yield correct versus incorrect answers. As shown in Figure~\ref{fig:density_gsm8k}, more than 90\% of rollouts with correct answers achieve a consistency score of at least 1, whereas this proportion drops to around 30\% for rollouts with incorrect answers. This sharp contrast indicates that correct rollouts are more robust across paraphrased variants, motivating the design of our paraphrase consistency mechanism to exploit this property for improved final predictions.

To address potential collisions when multiple answers obtain the same maximum consistency score, we introduce similarity-weighted consistency verification. Inspired by weighted majority voting~\citep{dogan2019weighted}, this approach adjusts the influence of each paraphrased form on the consistency score according to its similarity to the original form. Intuitively, paraphrased forms with lower similarity should exert greater weight, as they provide stronger evidence of robustness, whereas those closely resembling the original form contribute less. Concretely, we first extract embeddings for both the original and paraphrased questions and compute their similarity scores as $s_n = \text{sim}(q_0,q_n)$, where $\text{sim}(\cdot)$ denotes a similarity measure such as cosine similarity. We then derive weights via a softmax function $w_n = \text{softmax}(s_n)= \frac{\exp(-\lambda s_n)}{\sum_n \exp(-\lambda n)}$, where $\lambda$ is the temperature scaling coefficient. Finally, the similarity-weighted consistency score is defined as $\tilde{c}_k = \sum_{n-1}^N w_n \mathbb{I}(\Phi(r^{q_0}_k)=\Phi(r^{q_n}_k))$. This ensures agreement with more diverse paraphrases contributes more strongly to the final decision.

\begin{table}[t]
\centering
\caption{Comparison of our method with baseline approaches on Llama-3.1-8B-Instruct and Mistral-7B-Instruct-v0.2.}
\label{tab:main_1}
\centering
\resizebox{1.00\columnwidth}{!}{
\begin{tabular}{l|l|ccccc}
\toprule
\textbf{Model} & \textbf{Method} & GSM8K & GSM-Hard & Math500 & SVAMP & ARC \\
\hline
\multirow{7}{*}{Llama-3.1} 
& Chain-of-Thought & 77.40 & 28.00 & 31.00 & 83.00 & 58.91 \\
& Self-Consistency    & 80.60 & 31.80 & 37.80 & 85.10 & 60.75 \\
& Tree-of-Thought  & 75.74 & 33.28 & 31.60 & 81.20 & 80.72 \\
& Guided Decoding  & 75.51 & 32.45 & 31.20 & 81.70 & 81.74 \\
& Predictive Decoding  & 81.43 & 40.26 & 34.00 & 85.90 & 84.56 \\
& Phi-Decoding        & 86.58 & 39.88 & 38.20 & 84.50 & 85.41 \\
& PPCV~(Ours)          & \textbf{88.24} & \textbf{49.73} & \textbf{50.00} & \textbf{89.60} & \textbf{88.31} \\
\midrule
 \multirow{7}{*}{{Mistral-7B}} 
& Chain-of-Thought & 46.45 & 26.91 & 12.20 & 62.40 & 41.42 \\
& Self-Consistency & 50.38 & 28.65 & 14.20 & 66.70 & 44.54 \\
& Tree-of-Thought & 50.49 & 25.78 & 11.40 & 60.60 & 41.04 \\
& Guided Decoding & 50.79 & 27.07 & 14.00 & 62.90 & 39.51 \\
& Predictive Decoding & 55.67 & 27.07 & 14.40 & 62.10 & 47.87 \\
& Phi-Decoding & \textbf{56.60} & 28.43 & 13.40 & 63.20 & 60.24 \\
& PPCV~(Ours) & 56.58 & \textbf{31.08} & \textbf{14.60} & \textbf{69.30} & \textbf{69.88} \\
\bottomrule
\end{tabular}}
\vspace{-6mm}
\end{table}

\begin{table}[h]
\centering
\caption{Comparison of our method with baseline approaches on Qwen3-32B~(non-thinking mode).}
\label{tab:main_2}
\centering
\resizebox{1.00\columnwidth}{!}{
\begin{tabular}{l|l|cccc}
\toprule
\textbf{Model} & \textbf{Method} & AIME2024 & AIME2025 & BRUMO2025 & HMMT2025 \\
\hline
\multirow{5}{*}{Qwen3-32B} 
& Chain-of-Thought  & 30.00 & 23.67 & 30.00 & 9.67 \\
& Guided Decoding  & 26.67 & 22.67 & 28.67 & 7.33  \\
& Predictive Decoding  & 32.67 & 24.00 & 33.33 & 10.33  \\
& Phi-Decoding         & 33.60 & 24.33 & 36.67 & 10.67  \\
& PPCV~(Ours)          & \textbf{40.00} & \textbf{26.00} & \textbf{43.33} & \textbf{13.33}  \\
\bottomrule
\end{tabular}}
\vspace{-4mm}
\end{table}

\subsection{Discussion}
Our technical contributions differ from prior works in three distinct ways. First, prior works~\citep{zhou2024paraphrase,chen2024self,yadav2024pag} typically use paraphrasing merely to expand the solution space. In contrast, we introduce Paraphrastic Probing, a mechanism that uses paraphrasing to test the model's internal confidence. By analyzing the discrepancy in token-level logits of the initial trajectory between the original and paraphrased questions, we can rigorously pinpoint the critical tokens that may lead to errors in the following steps.This transforms paraphrasing from a generation tool into a precise, token-level diagnostic tool.
Second, prior works~\citep{zhou2024paraphrase,chen2024self} typically rely on simple majority voting across multiple solutions. Our Paraphrase Consistency mechanism is technically distinct. It validates answers based on their robustness across semantic variations of the problem constraint. We further introduce a similarity-weighted consistency metric that weighs answers based on the linguistic diversity of the paraphrase, offering a more nuanced selection criterion than simple frequency counts.
At last, a major technical limitation in current reasoning research is the reliance on external models or human-annotated error steps. Our method contributes a fully self-contained pipeline that identifies and corrects errors using the model's own sensitivity to surface-form perturbations. More discussion on the impact of critical tokens on correct trajectory can be found in Appendix~\ref{sec:app_dis}.

Besides, although we select the top candidate for the primary experiments to maintain computational efficiency, the framework itself naturally extends to the multi–critical-token setting. For multiple critical tokens, we can generate alternative tokens for each identified position and apply paraphrase consistency across the new rollouts. This allows the model to refine multiple segments of its intermediate reasoning steps rather than only one. The details of the algorithm can be found in Appendix~\ref{sec:app_algo}.

\begin{table}[t]
\centering
\caption{Comparison of model performance when using critical tokens versus random tokens.}
\label{tab:abla_1}
\centering
\resizebox{1.00\columnwidth}{!}{
\begin{tabular}{l|ccccc}
\toprule
 \textbf{Method} & GSM8K & GSM-Hard & Math500 & SVAMP & ARC \\
\hline
 Chain-of-Thought & 77.40 & 28.00 & 31.00 & 83.00 & 58.91 \\
random tokens        & 82.08 & 40.29 & 42.12 & 84.77 & 75.68 \\
critical tokens~(Ours)          & \textbf{88.24} & \textbf{49.73} & \textbf{50.00} & \textbf{89.60} & \textbf{88.31} \\
\bottomrule
\end{tabular}}
\vspace{-3mm}
\end{table}

\begin{table}[t]
\centering
\caption{Comparison of our proposed paraphrase consistency against the majority voting.}
\label{tab:abla_2}
\centering
\resizebox{1.00\columnwidth}{!}{
\begin{tabular}{l|ccccc}
\toprule
 \textbf{Method} & GSM8K & GSM-Hard & Math500 & SVAMP & ARC \\
\hline
 Chain-of-Thought & 77.40 & 28.00 & 31.00 & 83.00 & 58.91 \\
majority voting       & 87.20 & 47.36 & 48.19 & 88.80 & 86.16 \\
paraphrase consistency~(Ours)          & \textbf{88.24} & \textbf{49.73} & \textbf{50.00} & \textbf{89.60} & \textbf{88.31} \\
\bottomrule
\end{tabular}}
\vspace{-5mm}
\end{table}

\section{Experiments}
In this section, we first describe the experimental setup, followed by the main results of our proposed method compared to the baselines. We also perform ablation study and computational cost analysis.

\subsection{Setup}
\textbf{Datasets}. To comprehensively assess our method, we evaluate it on seven benchmarks. Six focus on mathematical reasoning, including GSM8K~\citep{cobbe2021training}, GSM-Hard~\citep{gao2023pal}, SVAMP~\citep{patel2021nlp}, Math500~\citep{hendrycks2measuring}, and the more challenging competition-level datasets AIME2024, AIME2025, BRUMO2025, and HMMT2025~\citep{balunovic2025matharena}. In addition, we use ARC-Challenge~\citep{clark2018think} to evaluate knowledge reasoning ability of LLMs.

\textbf{Baselines}. In our experiments, we use Chain-of-Thought~(CoT)~\citep{wei2022chain}, Self-Consistency~\citep{wangself}, Tree-of-Thought~(ToT)~\citep{yao2023tree}, Guided Decoding~\citep{xie2023self}, Predictive Decoding~\citep{manon}, and Phi-Decoding~\citep{xu2025phi} as baseline methods.

\textbf{Metric}. Following prior work, we adopt pass@k (k=1,4) as the primary evaluation metric.

\textbf{Implementation Details}. In our experiments, we adopt Llama-3.1-8B-Instruct~\citep{grattafiori2024llama}, Mistral-7B-Instruct-v0.2~\citep{Jiang2023Mistral7}, Qwen-3-32B~\citep{yang2025qwen3} and DeepSeek-R1-Distill-Llama-70B as the target models. we employ the non-thinking mode for Qwen-3-32B. Throughout our method, we employ the same model for generating paraphrased problems, identifying critical tokens, and producing new rollouts. In the first stage, we generate 4 paraphrased variants for each problem in the math benchmarks and 3 variants for each problem in the ARC dataset. In the second stage, we select the top 10 tokens for new rollouts, with the temperature scaling coefficient 
$\lambda$ set to 2. For fair comparison, we ensure a comparable inference budget across methods. Specifically, we rollout 48 samples for Self-Consistency~\citep{wangself}. For Predictive Decoding~\citep{manon} and Phi-Decoding~\citep{xu2025phi}, we rollout 4-8 samples per foresight step, and each problem typically involves 5–8 foresight steps. We also adopt a zero-shot CoT prompt to elicit the new rollouts. For the baselines, we strictly follow their original settings, including temperature values, sampling strategies, and the number of few-shot examples. All experiments are conducted on NVIDIA A100 GPUs.

\begin{figure*}[t]
\centering
\subfloat[]{%
  \includegraphics[width=0.28\textwidth]{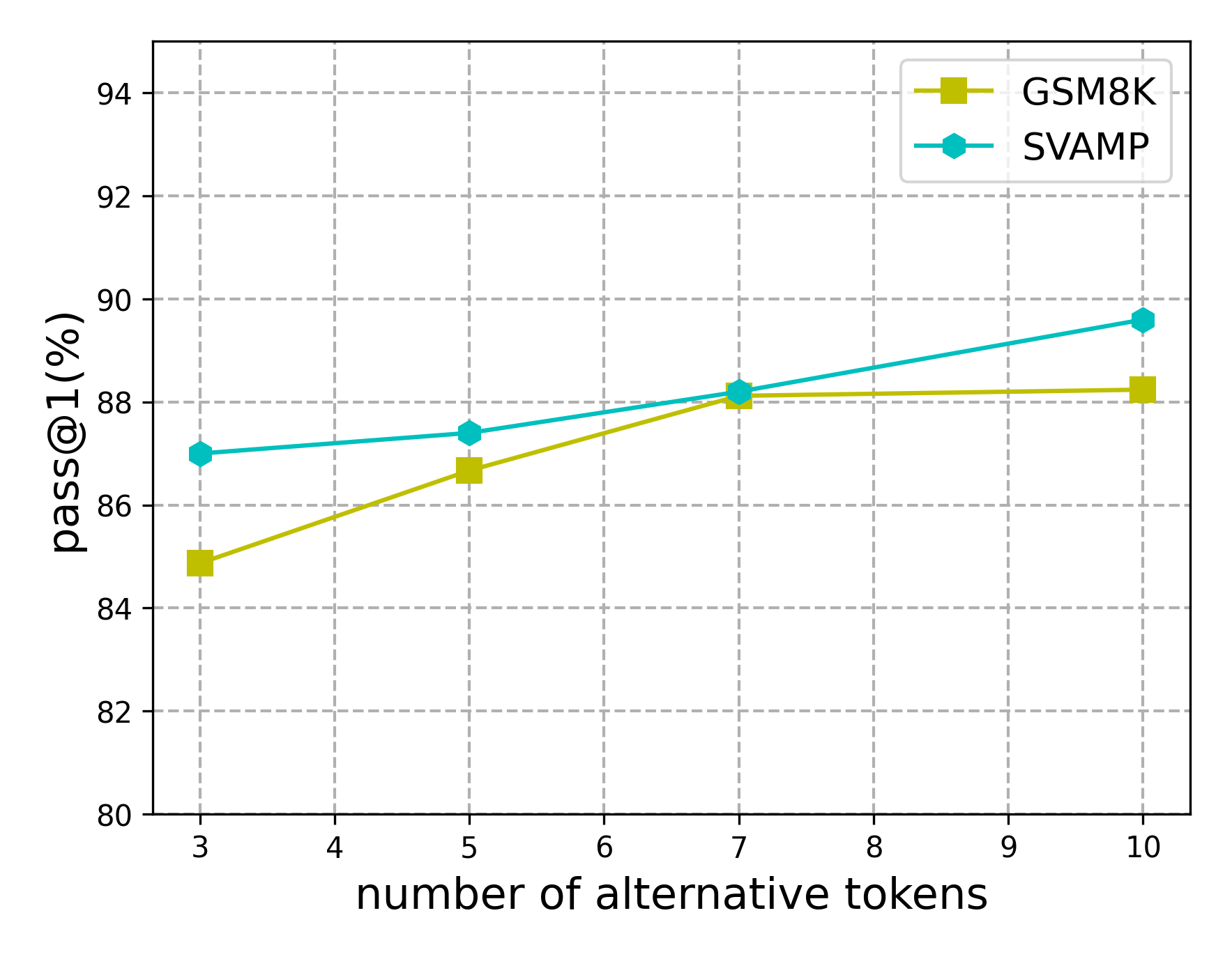}
  \label{fig:topk}
}
\subfloat[]{%
  \includegraphics[width=0.30\textwidth]{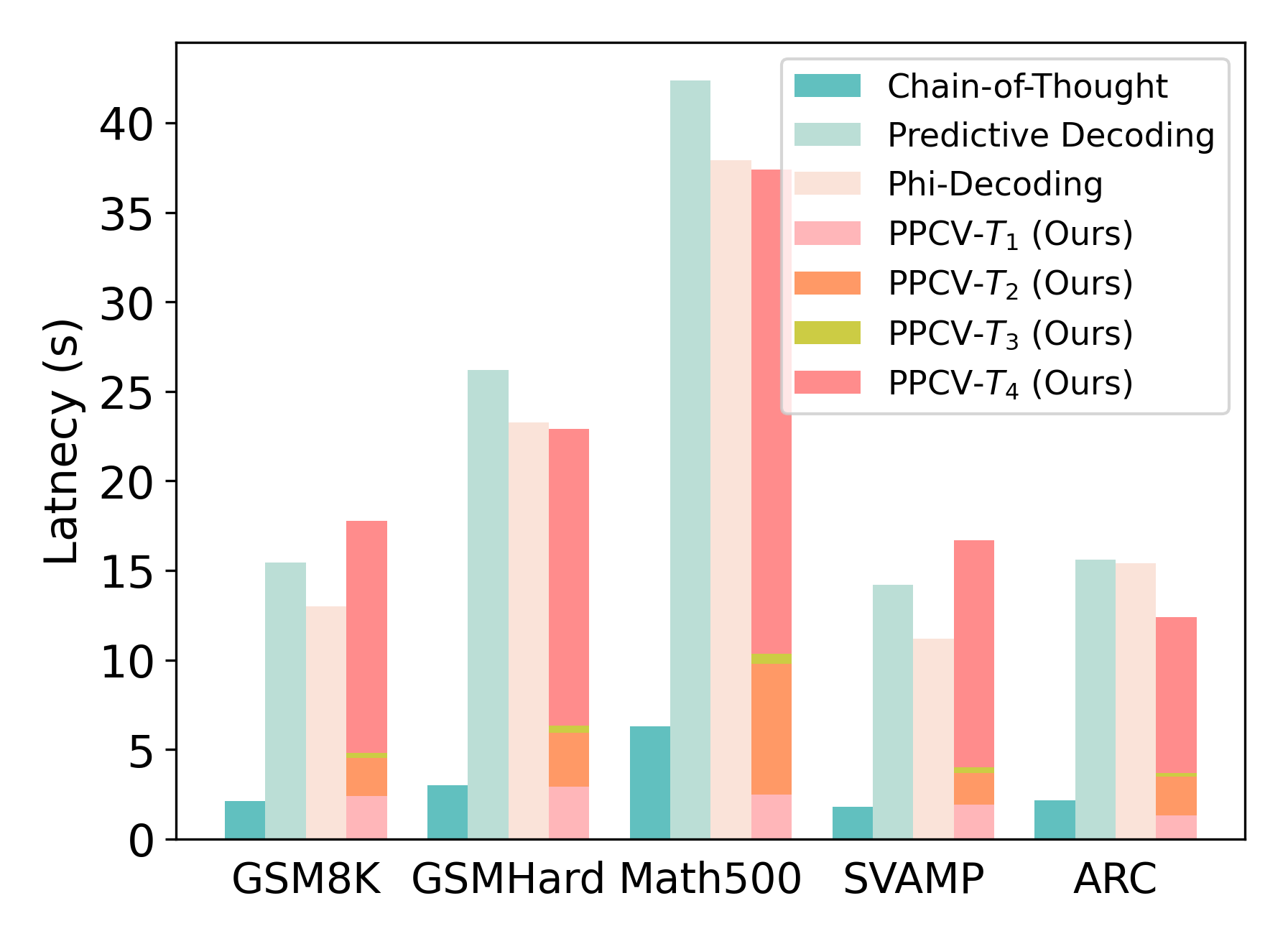}
  \label{fig:time}
}
\subfloat[]{%
  \includegraphics[width=0.30\textwidth]{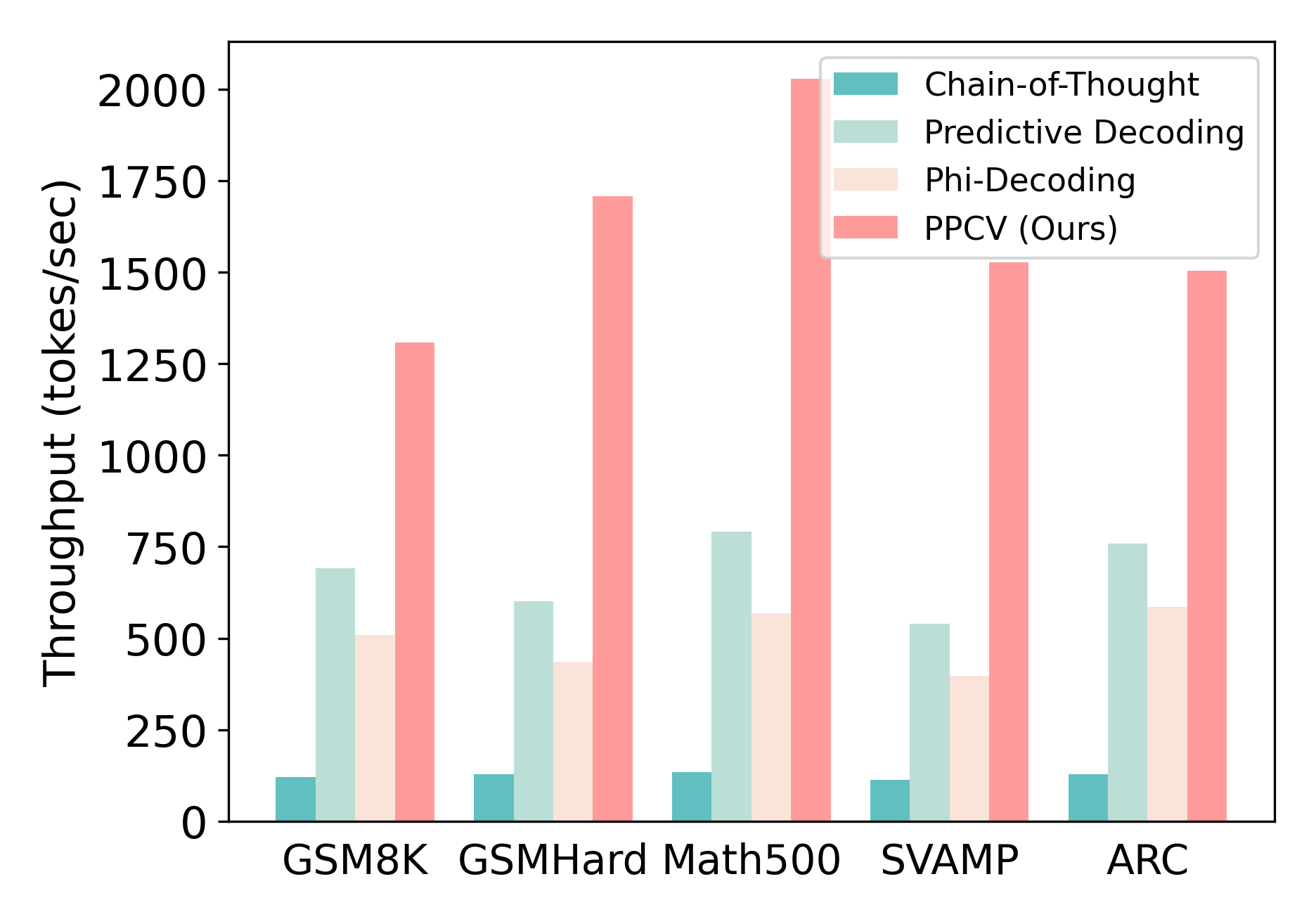}
  \label{fig:throughput}
}

\caption{(a) The impact of number of sampled alternative tokens on the performance. (b) Latency comparison between the baselines and our method, measured as the average inference time per question (in seconds). $T_1$, $T_2$, $T_3$, $T_4$ denote time for paraphrased question generation, initial answer generation, forward pass and new rollouts from truncated trajectories. (c) Throughput comparison between the baselines and our method, measured in output tokens per second.}
\vspace{-4mm}
\end{figure*}

\subsection{Main Results}
The main results are summarized in Table~\ref{tab:main_1} and Table~\ref{tab:main_2}. The results indicate that Self-Consistency effectively improves the reasoning performance of LLMs compared to vanilla Chain-of-Thought prompting. For example, Llama-3.1-8B-Instruct~\citep{grattafiori2024llama} achieves about 3\% higher accuracy with Self-Consistency than with CoT. These findings suggest that augmenting reasoning during inference through sampling is an effective way to refine reasoning trajectories. Recent decoding-based methods, such as Predictive Decoding~\citep{manon} and Phi-Decoding~\citep{xu2025phi}, also achieve strong results.Unlike prior works that rely on carefully designed prompts to self-correct errors in intermediate steps, these two methods modify the current step by probing future steps with pre-defined reward signals. Furthermore, our experimental results demonstrate that the proposed method consistently outperforms the baselines across most tasks, spanning both mathematical and knowledge reasoning, thereby highlighting its generalization ability across different reasoning settings. Notably, our method even surpasses the latest approaches such as Predictive Decoding~\citep{manon} and Phi-Decoding~\citep{xu2025phi}. In particular, it achieves approximately 50.00\% accuracy on the Math500 dataset~\citep{hendrycks2measuring}, exceeding these baselines considerably. The results on competition-level datasets further demonstrate the effectiveness of our method in enhancing the reasoning ability of LLMs. These results indicate that our method can effectively extract critical tokens that play a pivotal role in the final outcome and correct the reasoning trajectory by leveraging alternative tokens. Additional results can be found in Appendix~\ref{sec:app_results}.

\subsection{Ablation Study}
In this section, we analyze the contribution of each stage individually. All the evaluations are conducted on Llama-3.1-8B-Instruct~\citep{grattafiori2024llama}.

\textbf{Effectiveness of extracted critical tokens}. To demonstrate the effectiveness of our extracted critical tokens, we conduct an evaluation in which the critical tokens are replaced with random tokens in the first stage, while keeping the second stage unchanged. This evaluation is performed across multiple benchmark datasets, with pass@1 as the metric. The results, shown in Table~\ref{tab:abla_1}, reveal a substantial decline in performance. These findings highlight the pivotal role of critical tokens and indicate that our method can effectively identify and extract them. More ablation study on comparison with Paraphrased Majority Voting (PMV) can be found in Appendix~\ref{sec:app_abla}.

\textbf{Effectiveness of paraphrase consistency}. We also evaluate the effectiveness of our proposed paraphrase consistency and compare it with traditional majority voting. While keeping the first stage unchanged, instead of using paraphrased forms to generate new reasoning steps, we simply sample multiple new steps from alternative tokens conditioned on the original question and use majority voting to determine the final answer. The results, shown in Table~\ref{tab:abla_2}, reveal a noticeable decline in performance, highlighting the importance of paraphrased forms in improving the intermediate reasoning steps.

\textbf{Impact of number of sampled alternative tokens}.We investigate the influence of the number of sampled alternative tokens in the second stage by selecting values of 3, 5, 7, and 10. The results, shown in Figure~\ref{fig:topk}, demonstrate that performance improves as the number of alternative tokens increases. This suggests that exploring more reasoning steps with additional alternative tokens during inference can be beneficial for reasoning tasks.

\section{Computational Cost Analysis}

In this section we examine the composition of the latency in our method. The latency arises from four components: Paraphrased question generation~($T_1$); initial answer generation~($T_2$), equivalent to vanilla CoT; a forward pass for identifying critical tokens~($T_3$), which does not generate new tokens and is computationally lightweight; rollouts of truncated trajectories using alternative tokens under both the original and paraphrased questions~($T_4$), which constitutes the main source of overhead.

We evaluate all components on Llama-3.1-8B-Instruct using vLLM on NVIDIA A100 GPUs, with a maximum output length of 4096 tokens for each question. For our method, we use 4 paraphrased questions on math datasets and 3 on ARC, and select the top-10 candidate tokens as alternatives. The updated average latency results are reported in Figure~\ref{fig:time}. As expected, $T_1$ scales with the number of paraphrases, $T_3$ remains minimal, and $T_4$ dominates the total cost. Specifically, $T_4$ depends on the number of top-k alternative tokens, the number of paraphrased questions and the position of the critical token in the trajectory. Since the new rollouts from truncated trajectories for different alternative tokens and paraphrased questions are independent, $T_4$ can take advantage of vLLM’s parallelism. These rollouts can therefore be processed concurrently, improving overall efficiency. This is reflected in the higher throughput (tokens/sec) shown in Figure~\ref{fig:throughput}. And results of our method in latency comparable to baseline methods, even on challenging benchmarks such as Math500 and GSM-Hard where the critical token tends to occur in later reasoning steps. On the GSM8K and SVAMP benchmarks, our method, as well as baselines such as Predictive Decoding, would incur a approximately $6\text{–}8\times$ latency overhead compared to vanilla Chain-of-Thought. More analysis on the trade-off between the latency and performance  can be found in Appendix~\ref{sec:app_comp}.

\section{Conclusion}
In this study, inspired by beneficial impact of paraphrase forms on reasoning, we investigate the pivotal role of critical tokens in shaping the reasoning trajectory. To leverage these two factors, we propose the Paraphrastic Probing and Consistency Verification framework. Our framework consists of two stages: Paraphrastic Probing, which identifies and extracts critical tokens, and Consistency Verification, which adopts paraphrase forms to generate new reasoning trajectories with alternative tokens to reach the final answer. We evaluate our proposed framework with different LLMs and extensive evaluations across multiple benchmarks demonstrate the promising performance of our method.


\section*{Impact Statement}
This paper presents work whose goal is to advance the field of machine learning. There are many potential societal consequences of our work, none of which we feel must be specifically highlighted here.


\bibliography{reference}
\bibliographystyle{icml2026}

\newpage
\appendix
\onecolumn

\section{Empirical Study}
\label{sec:app_empirical}

\begin{figure*}[t]
\centering
\subfloat[]{%
  \includegraphics[width=0.25\textwidth]{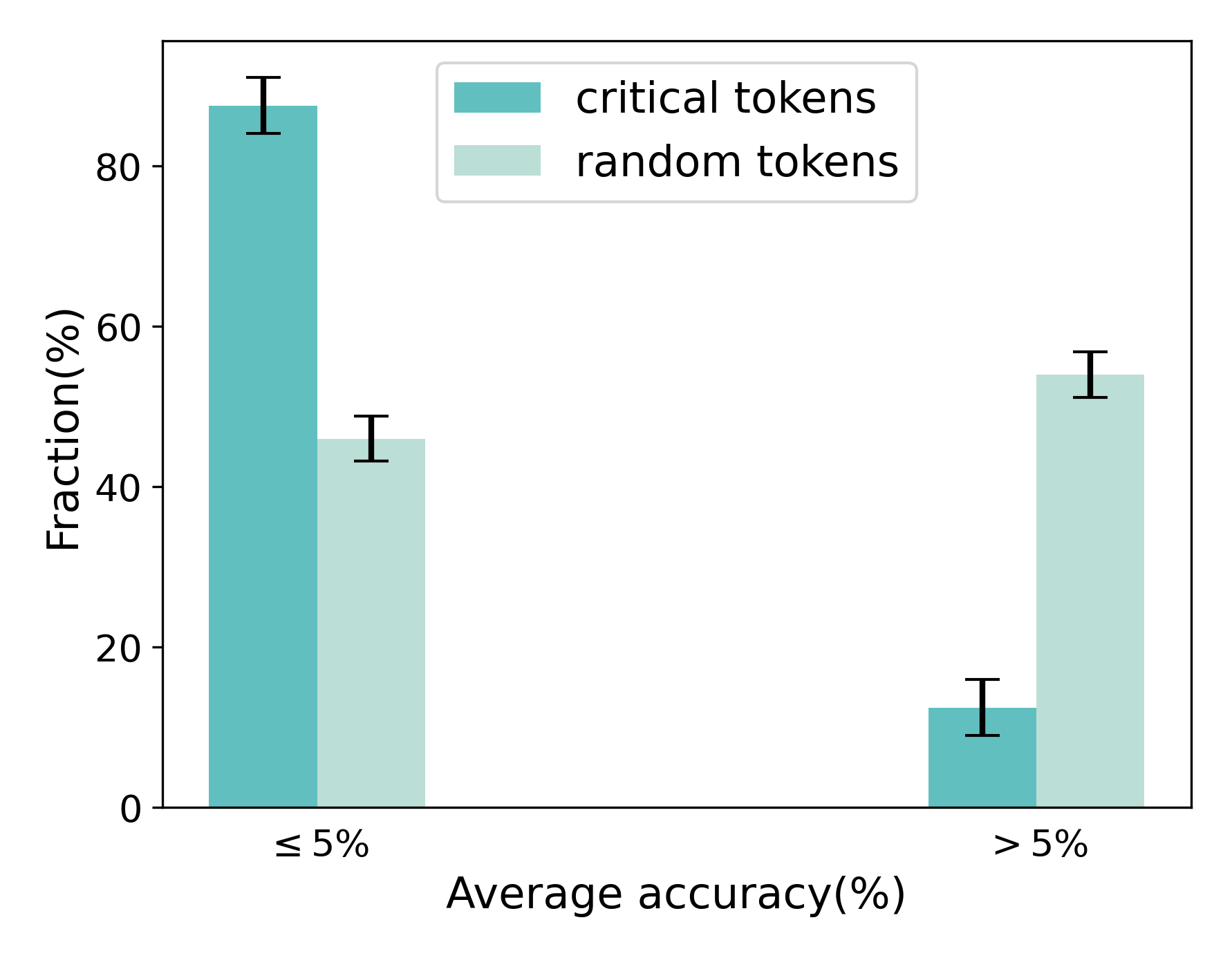}
  \label{fig:condition_one_math}
}
\subfloat[]{%
  \includegraphics[width=0.25\textwidth]{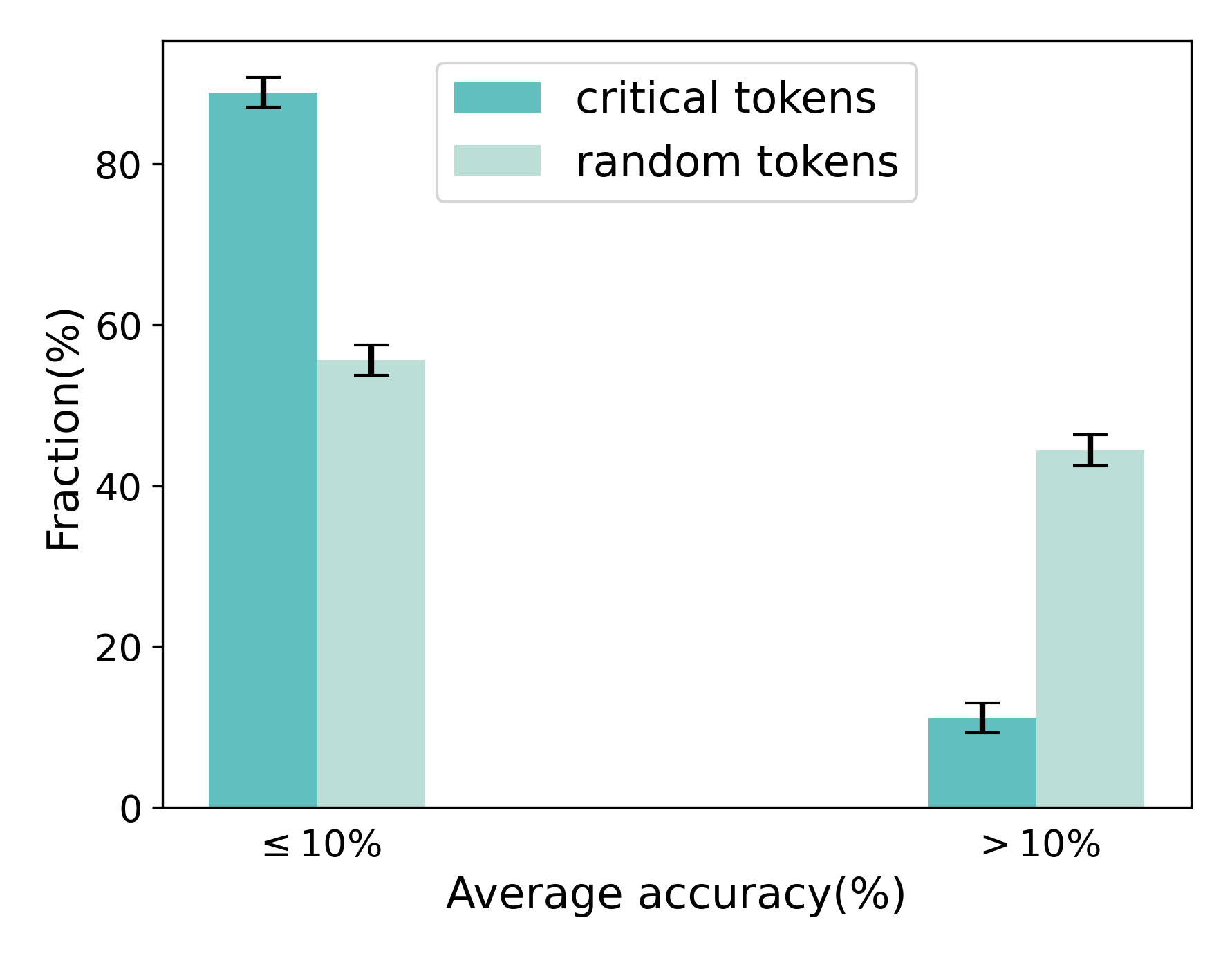}
  \label{fig:condition_two_math}
}
\subfloat[]{%
  \includegraphics[width=0.25\textwidth]{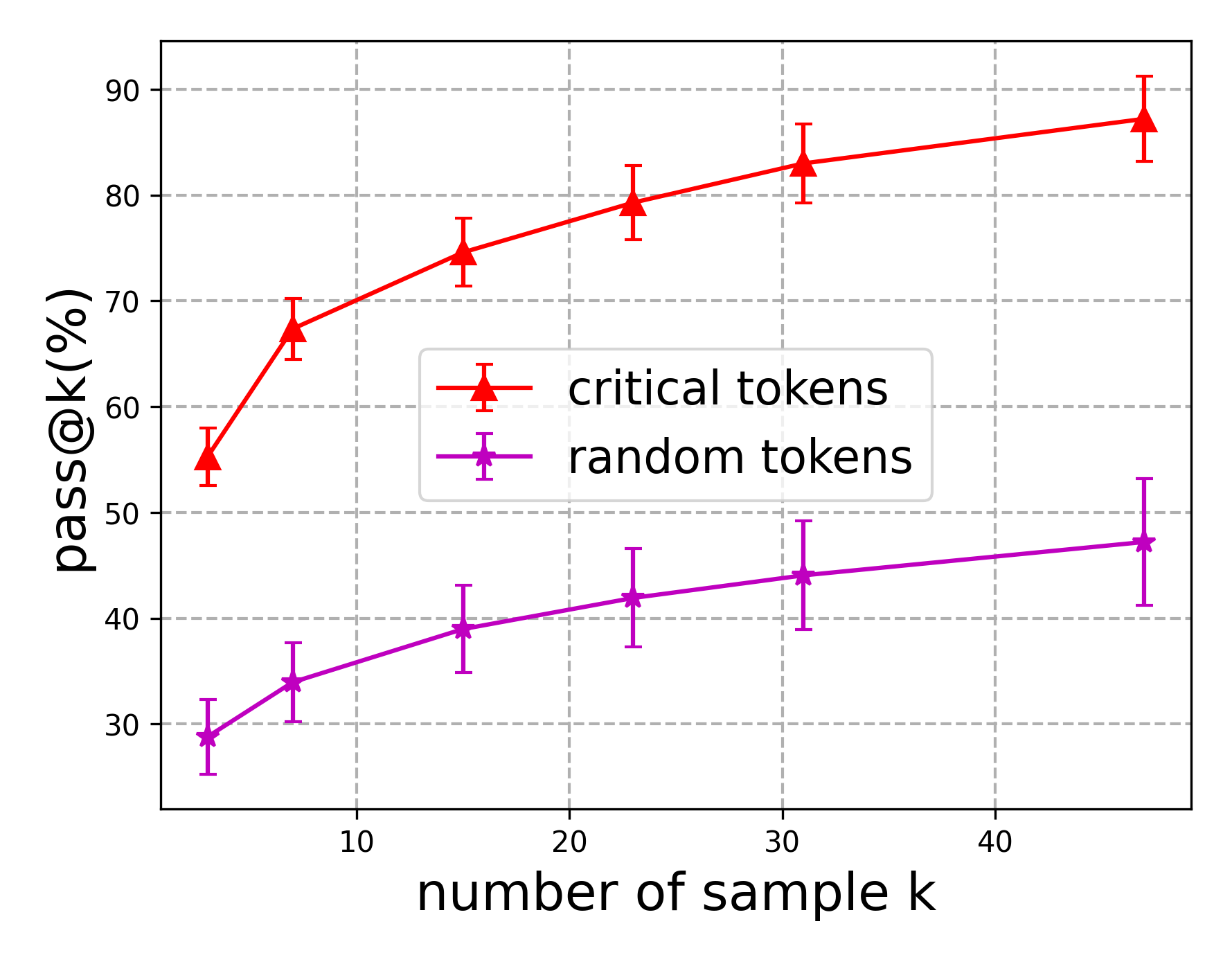}
  \label{fig:compare_ct_rnd_math}
}
\subfloat[]{%
  \includegraphics[width=0.25\textwidth]{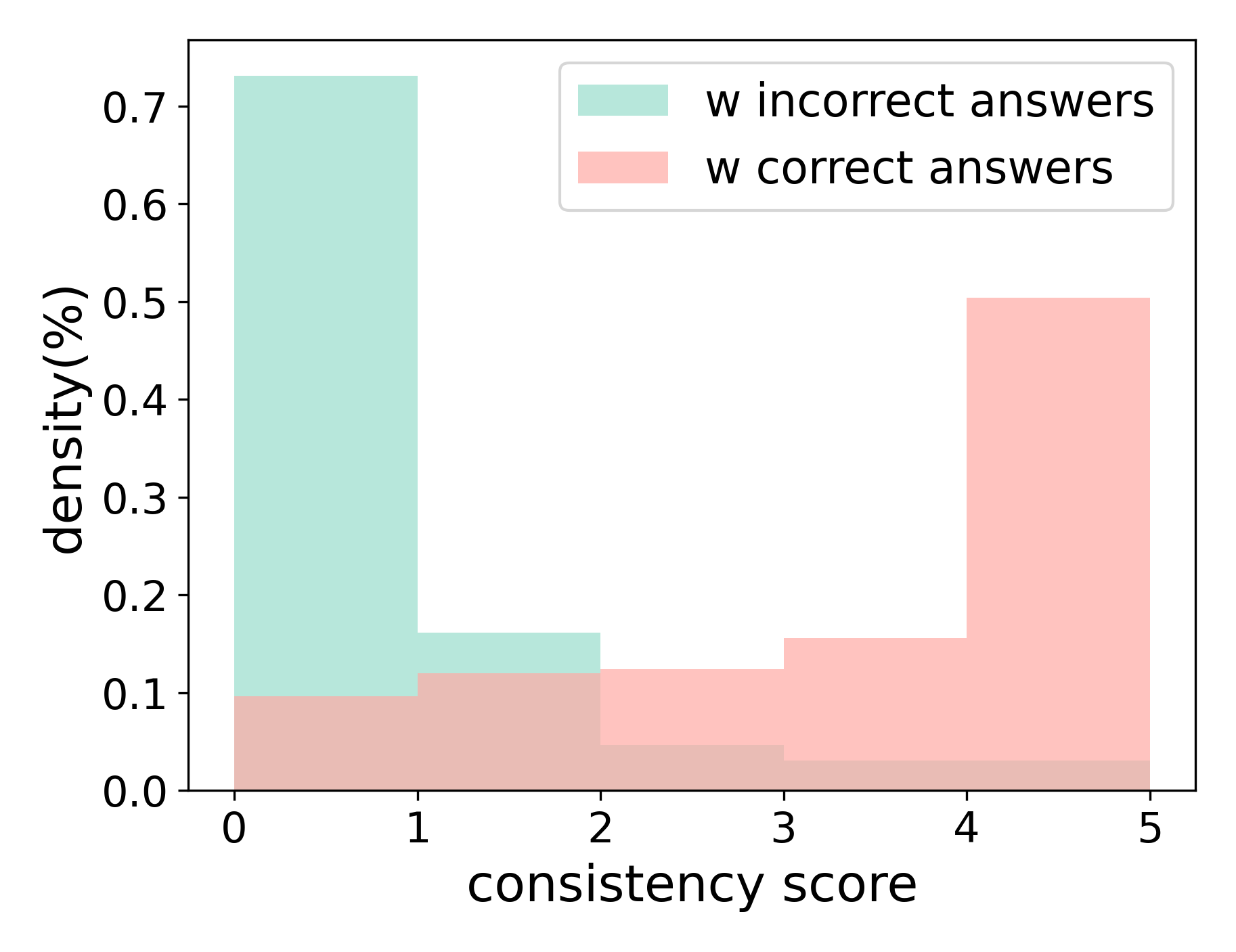}
  \label{fig:density_math}
}
\vspace{-2mm}
\caption{(a) Comparison of the fraction of tokens whose associated rollout correctness falls below or above 5\%, for tokens extracted by our method versus randomly selected tokens. (b) Comparison of the fraction of the subsequent tokens whose associated rollout correctness falls below or above 10\%, for tokens extracted by our method versus randomly selected tokens. (c) Comparison of the impact of extracted tokens by our method versus random tokens on LLM reasoning performance. (d) Comparison of the density distributions of consistency scores for rollouts with correct and incorrect answers. All the evaluations are conducted on Math500 dataset.}
\vspace{-4mm}
\end{figure*}

we follow the previous pipeline to conduct a quantitative analysis to examine the authenticity and impact of tokens extracted via paraphrastic probing, comparing them against randomly selected tokens. Specifically, we sample 100 instances with incorrect reasoning trajectories from the Math500~\citep{hendrycks2measuring} training sets. Following the paraphrastic probing pipeline, we identify candidate critical tokens in the early reasoning steps. In each run, we randomly sample 40\% of the candidate critical tokens for evaluation and repeat the experiment independently 10 times. For comparison, we apply the same procedure to randomly selected tokens. All evaluations are conducted using Llama-3.1-8B-Instruct~\citep{grattafiori2024llama}. We first compute the average accuracy of rollouts generated from partial trajectories truncated at the position of the extracted tokens, and compare these results with those obtained from random-token truncation. As shown in Figure~\ref{fig:condition_one_math}, we also obverse a large proportion of the extracted tokens exhibit near-zero correctness. We further evaluate the average accuracy of rollouts initiated from subsequent tokens (5–10 tokens). The results in Figure~\ref{fig:condition_two_math} also indicate that errors persist beyond the identified positions. Finally, we substitute the extracted critical tokens with alternative tokens and roll out multiple new reasoning trajectories. As shown in Figure~\ref{fig:compare_ct_rnd_math}, replacing critical tokens leads to a significantly larger improvement in reasoning accuracy compared to substituting random tokens.

To justify our paraphrase consistency mechanism on Math500 dataset~\cite{hendrycks2measuring}, we follow the same procedure. The evaluation is conducted on Llama-3.1-8B-Instruct~\citep{grattafiori2024llama}. We then analyze the distribution of consistency scores for rollouts that yield correct versus incorrect answers. As shown in Figure~\ref{fig:density_math}, majority of rollouts with correct answers achieve a consistency score of at least 1, whereas this proportion drops sharply for rollouts with incorrect answers.

\section{Automatic Prompt Engineering for Paraphrasing}
\label{sec:app_ape}

The quality of paraphrased questions is crucial for our framework, as it directly affects both the identification of critical tokens and the stability of paraphrase-based consistency verification. To ensure high-quality paraphrasing, we adopt Automatic Prompt Engineering~(APE)~\citep{zhou2022large} as an evaluation to optimize the paraphrasing instruction. This allows us to systematically control the quality of paraphrases rather than relying solely on manually written prompts. The APE procedure we employ consists of four steps:
\begin{itemize}
    \item  We create a small set of original questions paired with valid paraphrased variants. Using this exemplar set, the model generates multiple candidate instructions that could produce the paraphrased outputs from the original questions.
    \item Prompt each candidate instruction to the language model to generate paraphrases for the original problems and compare the mean solve rate change before and after paraphrasing.
    \item Choose the instruction that maximizes the mean solve rate change.
    \item Repeat the previous steps multiple times.
\end{itemize}
In our experiments, we initialize the paraphrasing instruction and iteratively refine it using APE with samples drawn from the GSM8K and Math500 training sets. With this approach, the refined paraphrasing prompt helps us reliably produce high-quality paraphrased questions.

\begin{algorithm}[t]
    \renewcommand{\algorithmicrequire}{\textbf{Input:}}
    \renewcommand{\algorithmicensure}{\textbf{Output:}}
    \caption{PPCV Framework}
    \label{alg:alg}
    \begin{algorithmic}
        \REQUIRE LLM; original question $q_0$; number of paraphrased questions $N$; number of sampled alternative tokens $K$; temperature-scaling coefficient $\lambda$.
        \ENSURE Final answer $\text{ans}_f$.

        \STATE \textbf{(Paraphrasing)} Generate paraphrased variants $q_1, q_2, \dots, q_N$ from $q_0$ using the LLM.

        \STATE \textbf{(Initial reasoning)} Obtain the initial reasoning path $r^{q_0}_0$ for $q_0$.

        \STATE \textbf{(Token influence analysis)}  
        For each paraphrased question $q_n$, compute the token distribution  
        $P^{q_n}$ by concatenating $q_n$ with $r^{q_0}_0$ (Eq.~\ref{eq:token_probability}).

        \STATE \textbf{(Candidate selection)}  
        Identify candidate critical token positions based on the mismatch between
        the top-1 predicted tokens and the expected tokens in $q_0$  
        (Eq.~\ref{eq:choose_candidate}).

        \STATE \textbf{(Critical token verification)}  
        Select the final critical token $a_c$ using the criteria  
        (Eqs.~\ref{eq:criterion_1}--\ref{eq:criterion_3}).

        \STATE \textbf{(Alternative token sampling)}  
        Sample $K$ alternative tokens  
        $a_c^0, a_c^1, \dots, a_c^{K-1}$  
        (including $a_c^0 = a_c$) using top-$K$ sampling on $q_0$.

        \STATE \textbf{(Truncated rollouts)}  
        Truncate the initial reasoning path $r^{q_0}_0$ at position $a_c$ and form synthetic inputs  
        $\tilde{r}_c^0, \tilde{r}_c^1, \dots, \tilde{r}_c^{K-1}$  
        by appending each alternative token.

        \FOR{each synthetic input $\tilde{r}_c^k$, $k = 0, \dots, K-1$}
            \STATE Generate rollouts  
            $r^{q_0}_k, r^{q_1}_k, \dots, r^{q_N}_k$  
            for the original and paraphrased questions using the LLM.
        \ENDFOR

        \STATE \textbf{(Consistency verification)}  
        Compute the final answer $\text{ans}_f$ using (similarity-weighted) paraphrase consistency  
        (Eq.~\ref{eq:consistency}).
    \end{algorithmic}
\end{algorithm}

\section{Algorithm}
\label{sec:app_algo}

The complete algorithm of our proposed PPCV framework is illustrated in Algo.~\ref{alg:alg}.

\section{Discussion}
\label{sec:app_dis}

\begin{wrapfigure}[16]{r}{0.5\textwidth}
  \vspace{-4mm}
  \includegraphics[width=0.5\textwidth]{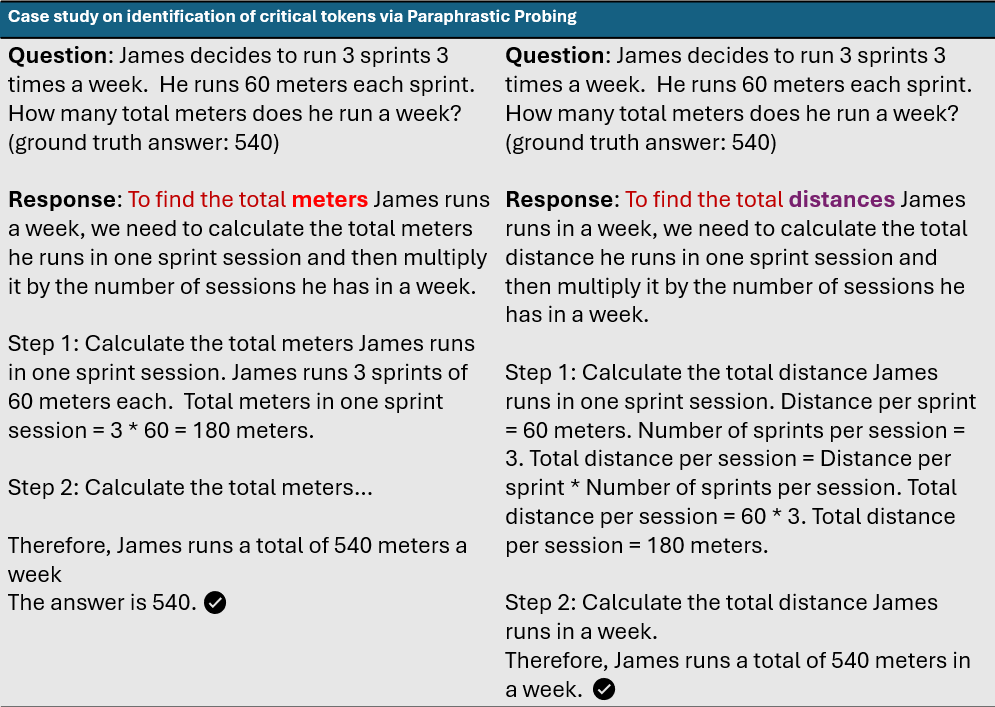}
    \caption{Effect of the critical token on an initial reasoning path that yields the correct answer. Critical tokens are highlighted in bold red, and alternative tokens in bold purple.}
    \label{fig:critical_token_on_correct}
\end{wrapfigure}
We have shown that replacing critical tokens can correct intermediate reasoning paths and lead to the correct answer. In this section, we examine how our method influences reasoning paths that are already correct. First, we conduct a case study on an instance with a correct answer, where we follow our pipeline to identify the critical token and roll out new reasoning paths using alternative tokens. As illustrated in Figure~\ref{fig:critical_token_on_correct}, the new rollouts also yield the correct answer. Second, our pipeline incorporates both the initial reasoning path $r^{q_0}_0$ and its paraphrased variants $r^{q_n}_0$ for evaluation. The robustness of correct rollouts across paraphrased forms ensures high consistency scores, allowing them to stand out as the final answer. These findings suggest that our pipeline preserves the performance of LLMs on problems that can already be solved correctly by CoT~\citep{wei2022chain}.

\begin{table}[t]
\centering
\caption{Comparison between our method and baselines with DeepSeek-R1-Distill-Llama-70B model on mathematical reasoning benchmarks using pass@1 accuracy.}
\label{tab:deepseek}
\centering
\resizebox{0.80\textwidth}{!}{
\begin{tabular}{l|cccc}
\toprule
\textbf{Method} & AIME2024 & AIME2025 & BRUMO2025 & HMMT2025 \\
\hline
Chain-of-Thought     & 56.67 & 38.00 & 43.33 & 30.00 \\
Predictive Decoding  & 60.00 & 40.66 & 44.66 & 30.66 \\
Phi-Decoding         & 64.00 & 46.67 & 48.00 & 31.33 \\
PPCV~(Ours)          & \textbf{70.00} & \textbf{56.66} & \textbf{56.66} & \textbf{33.33} \\
\bottomrule
\end{tabular}}
\end{table}

\begin{table}[t]
\centering
\caption{Comparison of our method and the baseline approach on Llama-3.1-8B-Instruct, evaluated using pass@k (k=4).}
\label{tab:passk_llama}
\centering
\resizebox{0.70\textwidth}{!}{
\begin{tabular}{l|ccccc}
\toprule
\textbf{Method} & GSM8K & GSM-Hard & Math500 & SVAMP & ARC \\
\hline
Phi-Decoding    & 92.15 & 53.57    & 52.60   & 91.19 & 90.73 \\
PPCV~(Ours)     & \textbf{93.83} & \textbf{61.41} & \textbf{59.39} & \textbf{94.48} & \textbf{94.24} \\
\bottomrule
\end{tabular}}
\end{table}

\begin{table}[!ht]
\centering
\caption{Comparison of our method and the baseline approach on Qwen3-32B, evaluated using pass@k (k=4).}
\label{tab:passk_qwen}
\centering
\resizebox{0.70\textwidth}{!}{
\begin{tabular}{l|cccc}
\toprule
\textbf{Method} & AIME2024 & AIME2025 & BRUMO2025 & HMMT2025 \\
\hline
Phi-Decoding    & 41.61 & 30.19 & 45.09 & 13.11 \\
PPCV~(Ours)     & \textbf{49.71} & \textbf{34.28} & \textbf{51.42} & \textbf{19.28} \\
\bottomrule
\end{tabular}}

\end{table}

\begin{table}[!ht]
\centering
\caption{The comparison of performance between Paraphrased Majority Voting (PMV) and our proposed PPCV using pass@1 accuracy.}
\label{tab:pmv}
\centering
\resizebox{0.70\textwidth}{!}{
\begin{tabular}{l|ccccc}
\toprule
\textbf{Method} & GSM8K & GSM-Hard & Math500 & SVAMP & ARC \\
\hline
 Chain-of-Thought & 77.40 & 28.00 & 31.00 & 83.00 & 58.91 \\
PMV             & 83.55 & 33.16    & 40.60   & 87.70 & 61.63 \\
PPCV~(Ours)     & \textbf{88.24} & \textbf{49.73} & \textbf{50.00} & \textbf{89.60} & \textbf{88.31} \\
\bottomrule
\end{tabular}}
\end{table}

\section{Main results}
\label{sec:app_results}

To further assess the effectiveness of our method, we evaluate it on a larger reasoning model such as DeepSeek-R1-Distill-Llama-70B. We apply our full pipeline and compare against all baselines across multiple mathematical reasoning benchmarks. The results, presented in Table~\ref{tab:deepseek}, show that our method consistently outperforms the baselines. These improvements demonstrate that our approach remains effective for stronger reasoning models and generalizes well beyond the smaller models.

we also conduct additional experiments to report pass@k performance for both our method and the baselines. In these experiments, we use Phi-Decoding as the representative baseline and evaluate on two models: Llama-3.1-8B-Instruct and Qwen3-32B. We set k= 4 and generate 12 samples per question to obtain stable estimates of pass@4.
The results across multiple benchmarks are presented in Table~\ref{tab:passk_llama} and Table~\ref{tab:passk_qwen}, respectively. Consistent with our main findings, our method achieves higher pass@k scores compared to the baseline methods, indicating that paraphrastic critical-token refinement continues to provide benefits in a multi-sample setting. These results further validate the robustness of our approach under stochastic sampling and demonstrate improved hit rates when multiple outputs are available.

\section{Ablation Study}
\label{sec:app_abla}

We also perform a comparison against a Paraphrased Majority Voting~(PMV) strategy is essential to demonstrate that the performance gains of our method are indeed derived from the critical token correction, rather than simply the benefit of having multiple paraphrased samples. Thus, we have performed the suggested additional experiment. We adopt Paraphrased Majority Voting~(PMV) as the baseline. For a fair comparison, this approach uses the same total number of trajectories generated from the paraphrased questions as PPCV, and reaches the final answer using standard majority voting over the rollouts from the paraphrased questions. Specifically, we use four variants of each question, with 10 rollouts per variant. We compare the PMV baseline against our full PPCV method and ensure the same number of paraphrased questions as in our method. The results shown in Table~\ref{tab:pmv} demonstrate that the performance yielded by PMV is significantly lower than that of our proposed PPCV method. This result confirms that simply rolling out trajectories from paraphrased questions and aggregating the answers is insufficient. The substantial performance gap between PMV and PPCV decisively highlights the necessity and effectiveness of our extracted critical tokens and the subsequent corrective reasoning step.

\begin{figure}[t]
\centering
\subfloat[]{%
  \includegraphics[width=0.40\textwidth]{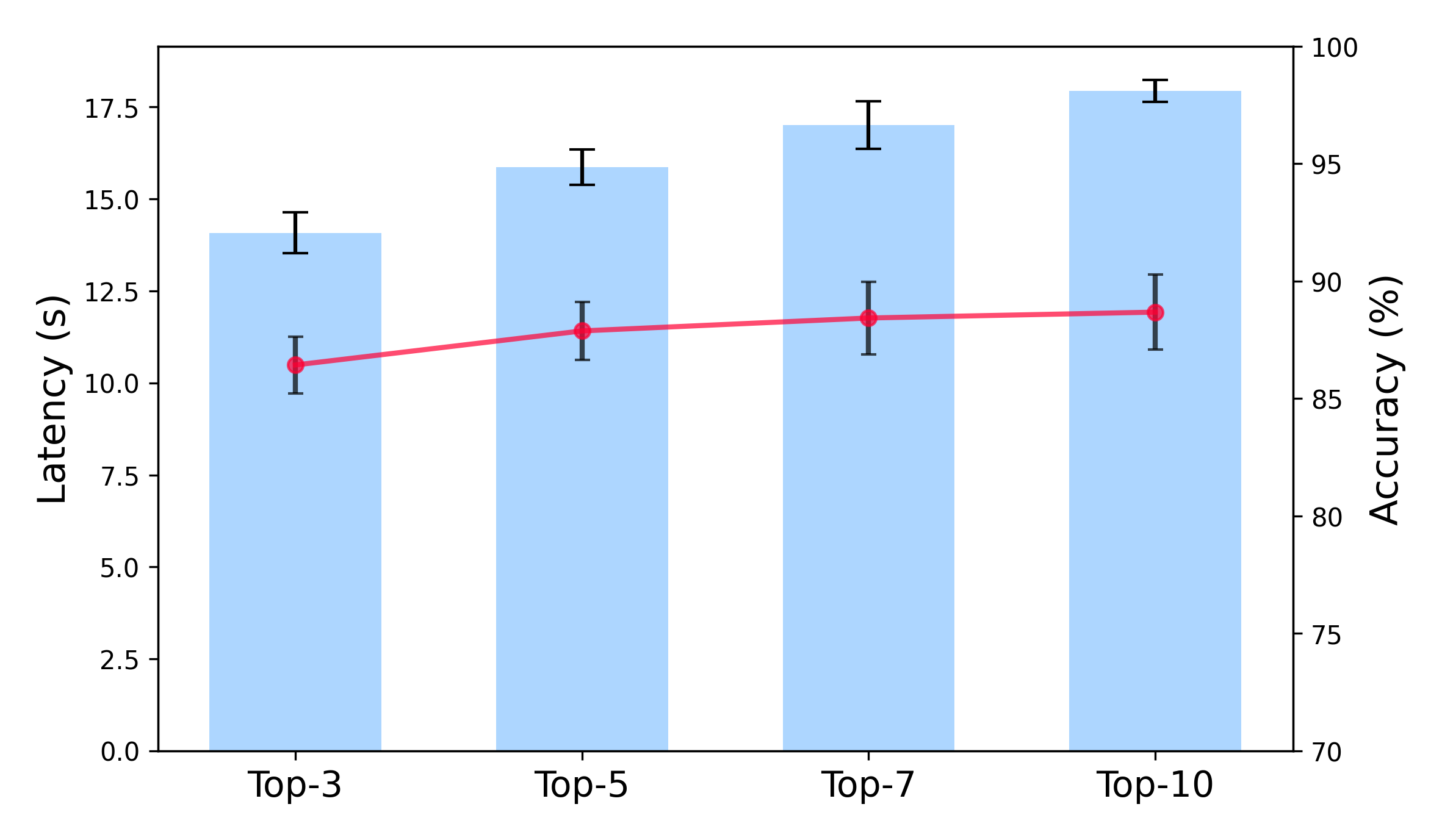}
}
\subfloat[]{%
  \includegraphics[width=0.40\textwidth]{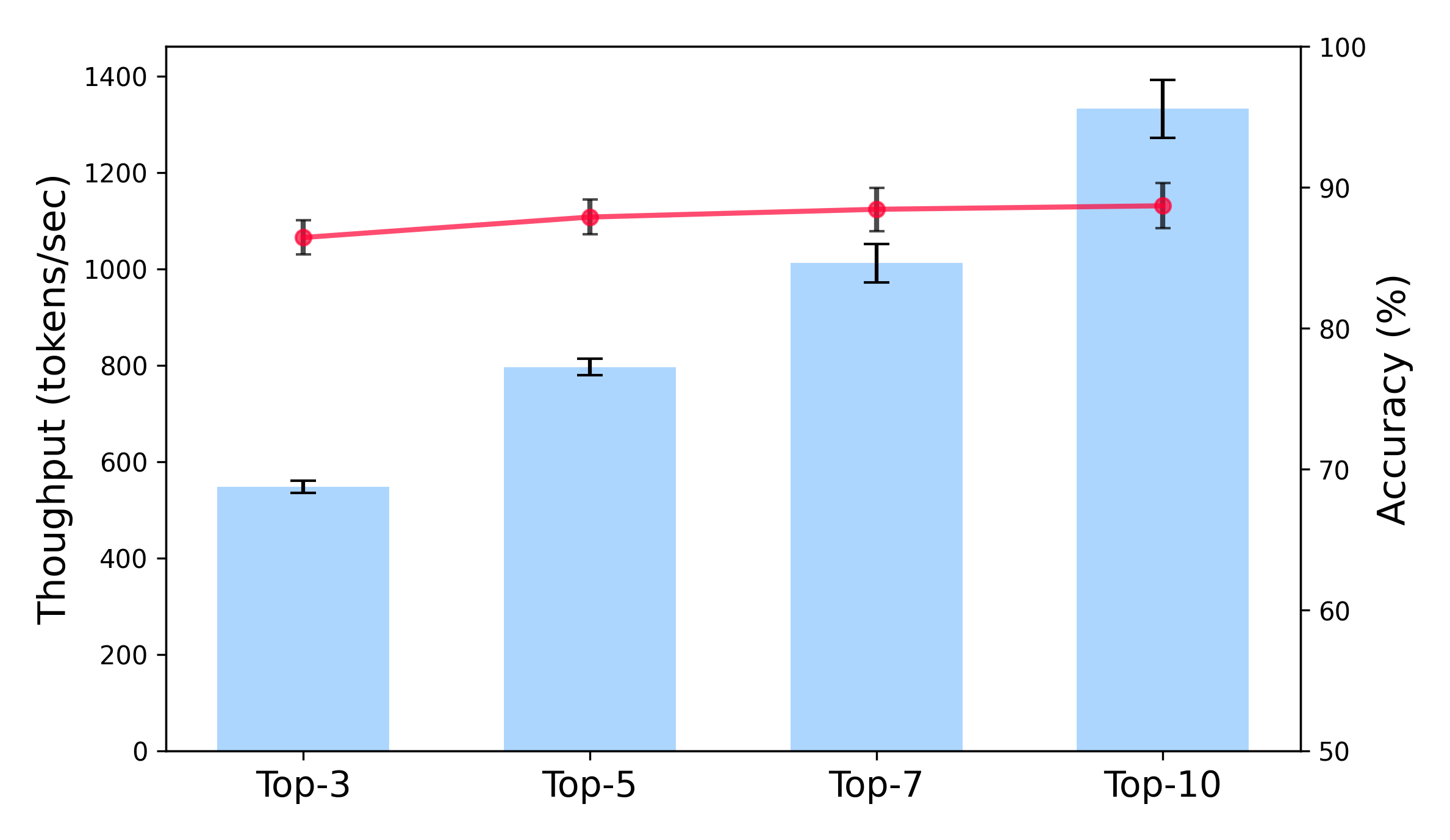}

}

\vspace{-2mm}
\caption{Trade-off between performance and efficiency for our method under different numbers of top-k tokens (k = 3, 5, 7, 10), measured by latency (left) and throughput (right).}
\label{fig:trade_off}
\vspace{-4mm}
\end{figure}

\section{Computational Cost Analysis}
\label{sec:app_comp}

We also conduct a trade-off analysis between performance and efficiency by varying the number of alternative tokens considered at the critical position. Specifically, we test top-3, top-5, top-7, and top-10 alternatives using GSM8K, sampling 200 random questions and repeating the experiment five times to compute confidence intervals. The results are presented in Figure~\ref{fig:trade_off}. We observe a clear trade-off. Reducing the number of alternative tokens lowers both latency and throughput, while causing a slight reduction in accuracy. This provides a practical mechanism for adjusting performance-efficiency trade-offs in real deployments. Depending on resource availability and target accuracy, practitioners can choose the appropriate number of alternative tokens.

\newpage

\subsection{Prompts}

\begin{tcolorbox}[
  colback=white,       
  colframe=black,      
  boxrule=1pt,         
  arc=2mm,             
  width=\textwidth,    
  sharp corners=south  
]
\textbf{Prompts for reasoning} (for math datasets)
\medskip

\textcolor{red}{System}: You are a specialized AI assistant designed to solve mathematical problems.

\medskip

\textcolor{red}{User}: Please solve the following problem step by step, and put your final answer within boxed\{\}.

\medskip

Each pack of dvds costs 76 dollars. If there is a discount of 25 dollars on each pack How much do you have to pay to buy each pack?
\medskip

\textcolor{red}{Assistant}: To find the price after the discount, we need to subtract the discount amount from the original price.

Step 1: Identify the original price and the discount amount.Original price = \$76. Discount amount = \$25

Step 2: Subtract the discount amount from the original price. Price after discount = Original price - Discount amount. Price after discount = \$76 - \$25. Price after discount = \$51

So, you have to pay \$51 to buy each pack.

boxed\{\$51\}

\end{tcolorbox}

\begin{tcolorbox}[
  colback=white,       
  colframe=black,      
  boxrule=1pt,         
  arc=2mm,             
  width=\textwidth,    
  sharp corners=south  
]
\textbf{Prompts for reasoning} (for ARC dataset)
\medskip

\textcolor{red}{System}: You are a specialized AI assistant designed to solve complex reasoning problems.

\medskip

\textcolor{red}{User}: Your task is to carefully read the following question, apply logical reasoning, scientific knowledge, and commonsense inference, and arrive at the correct answer. Always explain your reasoning step by step in clear language. At the end of your response, output the final choice in the following format: The answer is: (option), where option is the correct option letter (A, B, C, D, etc.), enclosed in parentheses.

\medskip

A group of engineers wanted to know how different building designs would respond during an earthquake. They made several models of buildings and tested each for its ability to withstand earthquake conditions. Which will most likely result from testing different building designs?

(A) buildings will be built faster

(B) buildings will be made safer

(C) building designs will look nicer

(D) building materials will be cheaper?
\medskip

\textcolor{red}{Assistant}: We are given a scenario in which engineers are testing different building designs to see how well they withstand earthquake conditions. The goal is to determine which of the listed outcomes is most likely to result from such testing.

Let 2019s evaluate each option in turn:

(A) Buildings will be built faster. 2013 This is not directly related to how well a building withstands earthquakes. Earthquake resistance is about structural integrity, not construction speed. While a good design might simplify construction, the primary purpose of the test is not speed. So this is not the most likely result.

(B) Buildings will be made safer 2013 This is directly aligned with the purpose of the experiment. Testing building designs under earthquake conditions helps identify which designs can better absorb or resist seismic forces. The goal is to improve safety during earthquakes. Therefore, this outcome is a direct and logical result of such testing.

(C) Building designs will look nicer 2013 Aesthetics are not the focus of earthquake testing. While a design might be both safe and attractive, the test itself does not evaluate visual appeal. So this is not a likely result of the testing.

(D) Building materials will be cheaper 2013 Cost of materials is not necessarily a result of testing for earthquake resistance. ...

The answer is: (B)
\end{tcolorbox}

\begin{tcolorbox}[
  colback=white,       
  colframe=black,      
  boxrule=1pt,         
  arc=2mm,             
  width=\textwidth,    
  sharp corners=south  
]
\textbf{Prompts to paraphrase the question} (for math datasets)
\medskip

\textcolor{red}{System}: You are a paraphrasing assistant specialized in math problems. Your task is to paraphrase a given math word problem. The core objective is to create a new, diverse expression of the problem while strictly preserving all numerical values, mathematical relationships, and the core logic required to solve it.

\medskip

\textcolor{red}{User}: Please create 5 paraphrased versions of the question below. Do NOT change any numbers or quantities. Maintain the exact mathematical operations and relationships. And vary the context and language. Please output paraphrased questions starting with a bracketed serial number like [1] [2] and ending with a new line.

\medskip

A robe takes 2 bolts of blue fiber and half that much white fiber.  How many bolts in total does it take?

\medskip

\textcolor{red}{Assistant}:

[1] Two bolts of blue fabric and one and a half times that amount of white fabric are required to make a robe. How many bolts in all are needed?
\medskip

[2] A robe is made from 2 bolts of blue yarn and an additional amount equal to half of that for white yarn. In total, how many bolts does it call for?

\medskip

[3] To create a robe, you'll need 2 bolts of blue thread and another half-quantity of white thread. How many bolts of thread in total should you purchase?

\medskip

[4] For a single robe, the production process demands 2 bolts of blue fabric and 1.5 times as much white fabric. What is the total number of bolts required?

\medskip

[5] The manufacture of a robe necessitates the usage of 2 bolts of blue material and slightly less than double that quantity for white material. How many total bolts does this equate to?

\end{tcolorbox}

\begin{tcolorbox}[
  colback=white,       
  colframe=black,      
  boxrule=1pt,         
  arc=2mm,             
  width=\textwidth,    
  sharp corners=south  
]
\textbf{Prompts to paraphrase the question} (for ARC dataset)
\medskip

\textcolor{red}{System}: You are given a multiple-choice question from a reasoning benchmark. 
Your task is to generate multiple different paraphrased versions of the question. 
Each version must preserve the original meaning, answer choices, and scientific/commonsense accuracy.

Guidelines:
- Keep the question logically equivalent to the original.
- Do not change the correct answer.
- Do not add or remove details that would alter the difficulty.
- Use different phrasing, sentence structures, or synonyms for each version.
- Keep the multiple-choice options (A, B, C, D, etc.) unchanged.

Example Input:
A large gene has dozens of exons, of which the central ones code for folded triple helical repeats that connect the cytoskeleton with sarcolemma and extracellular space. Each exon usually codes for one folded triple alpha helix. The most common mutations of the gene are central exon deletions that create out-of-frame peptides and progressive degenerative organ waste. A solution is to deliver a Morpholino that recognizes the 5' end of the out-of-frame exon in pre-mRNA. The molecule prevents binding of the spliceosome and creates exon skipping and in-frame joining. Several missing exons are well tolerated by an organism. Which structure below is not involved in the proposed therapy?

\medskip
(A) antisense
(B) lariat
(C) R-loops
(D) polyA tail

\medskip

\textcolor{red}{User}: Please create 3 paraphrased versions of the question below. Only paraphrases the question and leaves the multiple-choice options untouched. Please output paraphrased questions starting with a bracketed serial number like [1] [2] and ending with a new line,

\medskip
An astronomer observes that a planet rotates faster after a meteorite impact. Which is the most likely effect of this increase in rotation?

(A) Planetary density will decrease.

(B) Planetary years will become longer.

(C) Planetary days will become shorter.

(D) Planetary gravity will become stronger.

\medskip

\textcolor{red}{Assistant}:

[1] An astronomer notices that a planet's rotational speed increases following a meteorite collision. What is the most probable consequence of this accelerated rotation?  
(A) Planetary density will decrease.

(B) Planetary years will become longer.

(C) Planetary days will become shorter.

(D) Planetary gravity will become stronger.
\medskip

[2] After a meteorite strikes a planet, the astronomer observes an increase in its rotational velocity. What is the most likely outcome of this faster rotation? 
(A) Planetary density will decrease.

(B) Planetary years will become longer.

(C) Planetary days will become shorter.

(D) Planetary gravity will become stronger.

\medskip

[3] A planet begins to rotate more quickly after experiencing a meteorite impact. Which of the following is most likely to result from this increased spin? 
(A) Planetary density will decrease.

(B) Planetary years will become longer.

(C) Planetary days will become shorter.

(D) Planetary gravity will become stronger.

\end{tcolorbox}

\section{Use of Large Language Models}
In our work, we focus on improving large Malaguena models, and LLMs are used in our experimental evaluation. In addition, we employ LLMs to check grammatical errors and refine sentence structures.


\end{document}